\documentclass[lettersize,journal]{IEEEtran}

\usepackage[utf8]{inputenc}
\usepackage[T1]{fontenc}
\usepackage{amsmath,amsfonts,amssymb}
\usepackage{mathtools}
\usepackage{algorithmic}
\usepackage{algorithm}
\usepackage{array}
\usepackage{booktabs}
\usepackage{multirow}
\usepackage{rotating}
\usepackage{pdflscape}
\usepackage{textcomp}
\usepackage{stfloats}
\usepackage{url}
\usepackage{verbatim}
\usepackage{graphicx}
\usepackage{xcolor}
\usepackage{colortbl}
\usepackage{diagbox}
\usepackage{nicefrac}
\usepackage{microtype}
\usepackage{enumitem}
\usepackage{lipsum}
\usepackage{tabularx}
\usepackage{ragged2e}
\usepackage{adjustbox}
\usepackage[normalem]{ulem}
\usepackage{bm}
\usepackage{cite}
\usepackage[hidelinks]{hyperref}

\usepackage{amsmath,amsfonts,bm}

\def\eqref#1{equation~\ref{#1}}
\def\1{\bm{1}}

\DeclareMathAlphabet{\mathsfit}{\encodingdefault}{\sfdefault}{m}{sl}
\SetMathAlphabet{\mathsfit}{bold}{\encodingdefault}{\sfdefault}{bx}{n}

\useunder{\uline}{\ul}{}

\newcommand{\citep}[1]{\cite{#1}}
\newcommand{\citet}[1]{\cite{#1}}
\providecommand{\ContinuedFloat}{}

\setlist{nosep,leftmargin=*}

\begin{document}
\bstctlcite{IEEEtranBSTCTL}

\title{PhyxMamba: Chaotic System Reconstruction from Short Context Observations with Generative State-Space Models}

\author{Chang~Liu, Bohao~Zhao, Jingtao~Ding, Huandong~Wang, and Yong~Li%
\thanks{Chang Liu, Bohao Zhao, Jingtao Ding, Huandong Wang, and Yong Li are with the Department of Electronic Engineering, BNRist, Tsinghua University, Beijing, China (e-mail: lc23@mails.tsinghua.edu.cn; zhao-bh23@mails.tsinghua.edu.cn; dingjt15@tsinghua.org.cn; wanghuandong@tsinghua.edu.cn; liyong07@tsinghua.edu.cn).}}

\maketitle

\begin{abstract}
  Understanding chaotic dynamics is a fundamental problem across scientific disciplines, including climate science, neuroscience, and fluid dynamics, yet direct experimentation and intervention in such systems are often infeasible. Chaotic system reconstruction aims to identify a surrogate dynamical model that preserves a system’s invariant geometric and long-term temporal signatures from observed time series, thereby providing a controllable foundation for probing its mechanisms through systematic perturbation and analysis. However, faithful system reconstruction is hampered by high observational costs, which often restrict data to short, discontinuous sequences spanning only limited timescales. Conventional approaches such as reservoir computing struggle in this data-scarce regime since they typically require long-term synchronization windows to localize states on the attractor. Similarly, while deep learning-based time-series forecasting models effectively fit local trajectories, they often fail to capture global invariants, leading to a collapse of long-term dynamical integrity. Here, we propose PhyxMamba, a framework that synergizes Mamba-based state-space models with physics-informed principles. By leveraging time-delay embeddings to reconstruct the attractor manifold and employing a generative training scheme with geometry-aware regularization, PhyxMamba effectively captures both fine-grained local evolution and global physical constraints. Extensive experiments on simulated and real-world chaotic systems demonstrate that PhyxMamba achieves superior reconstruction performance, outperforming the strongest baseline by over 44\% in prediction accuracy and 8\% in topological fidelity on the Lorenz96 system, while exhibiting strong robustness against partial observations and noise. Codes are available at \url{https://github.com/changliu01/PhyxMamba}.

\end{abstract}


\section{Introduction} \label{sec:intro}
Chaotic systems, defined by their deterministic nature and extreme sensitivity to initial conditions, are pervasive across a wide range of scientific and engineering disciplines, including climate science~\cite{shukla1998predictability}, epidemiology~\cite{jones2021spread}, neuroscience~\cite{vignesh2025review}, and financial modeling~\cite{vogl2022controversy}. However, rigorous inquiry into these complex dynamics is often impeded by the infeasibility of direct experimentation; for instance, conducting controlled trials on climate systems or monitoring sensitive biological networks without disrupting them is infeasible.
Consequently, learning a faithful surrogate model from observational data, known as chaotic system reconstruction, is essential. Unlike forecasting that prioritizes point-wise accuracy, reconstruction seeks to preserve global topological invariants (e.g., fractal dimensions, state-space distribution) to enable valid long-term simulations. However, in many applications, models must infer the current dynamical state and initiate autonomous rollouts from short context observations available at inference time or within each training segment. Such contexts may span only limited characteristic timescales (e.g., Lyapunov times), making it difficult to recover the global attractor geometry required for faithful reconstruction.

Existing paradigms for chaotic system reconstruction range from classical analytical models defined by differential equations~\cite{lorenz1996predictability,lorenz2017deterministic,rossler1976equation} to contemporary data-driven approaches~\cite{pathak2018model,mikhaeil2022difficulty,hess2023generalized}.
Analytical models are rendered impractical in real-world reconstruction scenarios where the governing equations are unknown or the system parameters are unobservable.
Conversely, data-driven methods seek to infer dynamics directly from observations. Recurrent sequence models, such as Reservoir Computing (RC)~\cite{pathak2018model,gauthier2021next,li2024higher} and Piecewise-Linear Recurrent Neural Network (PLRNN)~\cite{mikhaeil2022difficulty,hess2023generalized}, rely heavily on the synchronization of internal reservoir states or specialized latent structures that necessitate extended continuous observation windows. Deep time-series predictors based on MLPs and Transformers~\cite{zhang2023crossformer,liu2023itransformer,ekambaram2023tsmixer,leguen2023deep, chien2023continuous, liu2024gph} primarily optimize point-wise trajectory accuracy, but they do not explicitly preserve a chaotic system's invariant measure. More recently, state-space models represented by Mamba~\cite{gu2023mamba} have emerged as promising candidates because their continuous-time latent states provide a favorable inductive bias for physical evolution. Yet using state-space models as generic forecasters is still insufficient for chaotic reconstruction: the model should be generative after a short context initialization and must preserve invariant statistics under long autonomous rollouts. How to build such generative state-space models for chaotic system reconstruction from short context observations remains largely underexplored.

The key challenge lies in the intrinsic tension between limited observational context and chaotic complexity, which gives rise to three critical issues: (i) \textit{\textbf{Extracting complex dynamical information from short context windows.}} It requires the model to infer the governing laws of the system by deciphering the latent topological information embedded within brief evolutionary segments. (ii) \textit{\textbf{Mitigating error amplification during autoregressive evolution.}} Driven by positive Lyapunov exponents, microscopic approximation errors accumulate exponentially over time. An effective generative modeling process is essential to suppress this intrinsic divergence and support stable autonomous rollout. (iii) \textit{\textbf{Enforcing ergodic invariance and geometric fidelity.}} It lies in constraining the model evolution within the manifold of the strange attractor, thereby precluding dynamical collapse into trivial states.

To address these challenges, we propose PhyxMamba, a framework that unifies a Mamba-based architecture with geometric regularization principles to reconstruct the underlying dynamics of chaotic systems. First, we introduce a physics-informed patch representation via time-delay embedding. It synthesizes coupled multivariate dynamics to reconstruct local phase-space states from short contexts, thereby equipping the model with essential geometric priors. Second, we design a multi-step aware generative training strategy based on the Mamba backbone~\cite{gu2023mamba,DBLP:conf/icml/DaoG24}. By incorporating multi-horizon perception into the generative process, it mitigates error accumulation and improves autoregressive stability. Third, we implement an invariant measure alignment stage. Utilizing a student-forcing strategy regularized by Maximum Mean Discrepancy (MMD)~\cite{li2017mmd}, we enforce statistical consistency between generated trajectories and the true system dynamics, preventing geometrical collapse. Our contributions are summarized as follows:

\begin{itemize}[leftmargin=*,partopsep=0pt,topsep=0pt]

\item \textbf{A generative state-space framework for chaotic system reconstruction.} We propose PhyxMamba, a framework that integrates the modeling capabilities of the Mamba backbone with physics-informed representations and regularizations. It enables robust reconstruction of chaotic systems from short context observations.

\item \textbf{Tailored architectural designs for reconstruction fidelity.} The model components unify representation, evolution, and regularization. It infers local phase-space structure from short context windows, stabilizes long-horizon generative rollouts via internalized multi-step evolution guidance, and simultaneously enforces explicit constraints on the system's invariant measure. These designs ensure that the learned model evolves strictly on the correct attractor manifold.

\item \textbf{Comprehensive empirical validation of performance and robustness.} Extensive evaluations across diverse simulated and real-world datasets demonstrate that PhyxMamba consistently achieves state-of-the-art reconstruction quality. Crucially, our framework remains robust to non-ideal conditions, maintaining effectiveness even under partial observations, varying sampling resolutions, and observational noise.
\end{itemize}

\section{Preliminaries and Problem Formulation} \label{sec:pfp}
In this section, we start with the theoretical preliminaries and notations, then formally define the chaotic system reconstruction problem.

\subsection{Chaotic System}
A chaotic system is characterized by deterministic evolution exhibiting extreme sensitivity to initial conditions, where infinitesimal perturbations lead to exponentially diverging trajectories in phase space. Formally, consider a continuous-time dynamical system governed by the autonomous ordinary differential equation (ODE):
\begin{equation}
    \frac{d\bm{x}(t)}{dt}=\bm{f}(\bm{x}(t)),
\end{equation}
where $\bm{x}(t) \in \mathbb{R}^V$ represents the $V$-dimensional system state at time $t$, and $\bm{f}:\mathbb{R}^V \rightarrow \mathbb{R}^V$ denotes a smooth nonlinear vector field driving the dynamics. A defining hallmark of such systems is the existence of a strange attractor, a compact invariant set $\mathcal{A} \in \mathbb{R}^V$ to which trajectories asymptotically converge. Unlike simple fixed points or limit cycles, the strange attractor possesses a complex fractal geometry~\cite{theiler1990estimating} and is supported by an invariant measure~\cite{jiang2023training,schiff2024dyslim}, describing the system's ergodic statistical properties.

\subsection{Lyapunov Exponent and Lyapunov Time}
Lyapunov exponents (LEs) quantify the average exponential rate of divergence or convergence of initially nearby trajectories in phase space, serving as the fundamental measure of a system's sensitivity to initial conditions. Formally, for two trajectories separated by an infinitesimal perturbation vector $\bm{\delta}_0$, their divergence over time $t$ evolves as:
\begin{equation}
    |\bm{\delta}(t)| \approx e^{\lambda t}|\bm{\delta}_0|.
\end{equation}
A $V$-dimensional dynamical system possesses a spectrum of $V$ Lyapunov exponents, $\{\lambda_1, \lambda_2, \dots, \lambda_V\}$, sorted in descending order. Among these, the Maximal Lyapunov Exponent (MLE), denoted as $\lambda_{\text{max}}$, dominates the long-term separation rate. A positive MLE ($\lambda_{\text{max}} > 0$) is the defining signature of chaos. 
Given the directional dependence of the separation vector, the Lyapunov exponent exhibits variable values. Among these, we focus on the maximal Lyapunov exponent (MLE). Typically, a positive MLE indicates that the system is chaotic. The MLE is mathematically defined as:
\begin{equation}
    \lambda_{\text{max}} = \lim_{t \rightarrow \infty} \lim_{|\bm{\delta}_0|\rightarrow 0} \frac{1}{t}\ln \frac{|\bm{\delta}(t)|}{|\bm{\delta}_0|},
\end{equation}
where $|\bm{\delta}_0|\rightarrow 0$ guarantees the applicability of the linear approximation across all times. Associated with the MLE is the Lyapunov Time, i.e., $T_L = {1}/{\lambda_{\text{max}}}$. This quantity represents the characteristic timescale for reliable trajectory prediction, marking the horizon where microscopic uncertainties amplify to macroscopic scales, making precise forecasting a significant challenge.

\subsection{Chaotic System Reconstruction}
Consider a continuous-time dynamical system with $V$ variables.
Given a collection of short observation-target segments
\begin{equation}
    \mathcal{D}_{\mathrm{train}}=\{(\bm{X}^{\mathrm{obs}}_i,\bm{X}^{\mathrm{tar}}_i)\}_{i=1}^{N},
\end{equation}
where $\bm{X}^{\mathrm{obs}}_i \in \mathbb{R}^{V \times T}$ contains only a short context, and $\bm{X}^{\mathrm{tar}}_i$ denotes the corresponding future segment used to train the model's evolution. Our goal is to learn an autonomous generative model $\mathcal{M}_\theta$ that can roll out future states over a much longer horizon $H \gg T$. At test time, the model receives only an unseen short context $\bm{X}^{\mathrm{obs}}_* \in \mathbb{R}^{V \times T}$ and recursively generates $\hat{\bm{X}}_{T+1:T+H} \in \mathbb{R}^{V \times H}$. The objective is not only to match the near-term trajectory within the predictable horizon, but also to preserve the invariant statistics and attractor geometry of the underlying system during long-term autonomous rollout. The specific evaluation metrics are detailed in Section~\ref{sec:exp-setup}.

\section{Methods} \label{sec:methods}
\begin{figure*}[t]
    \centering
    \includegraphics[width=\textwidth]{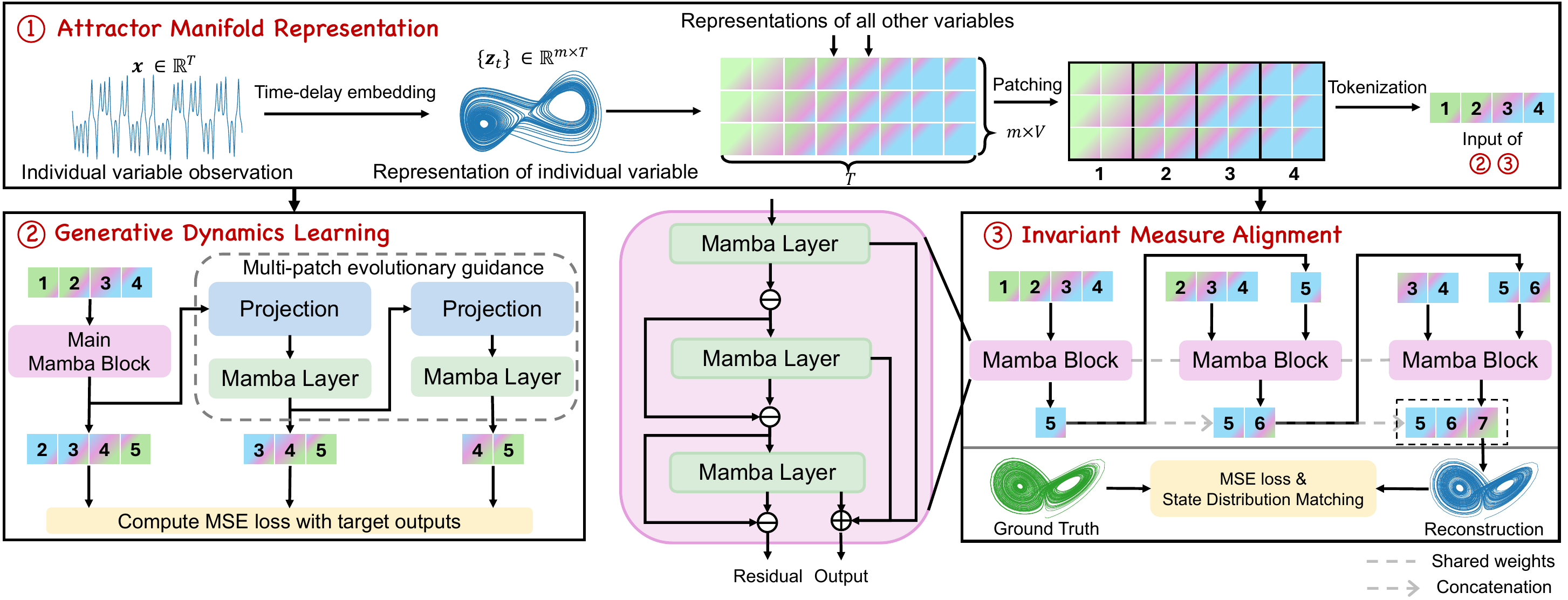}
    \caption{Overview of the proposed framework PhyxMamba.}
    \label{fig:model}
\end{figure*}

% Overview
% Step1：嵌入 + Patch
% Step2：Multi-token prediction，teacher & Student forcing
% Step3：Loss：MSE + MMD
In this section, we introduce PhyxMamba, a generative state-space framework that reconstructs chaotic systems from short context observations. The overall architecture, shown in Figure~\ref{fig:model}, comprises three components aligned with the challenges above: (i) attractor manifold representation, which uses physics-informed time-delay embeddings to recover local phase-space structure from short contexts; (ii) generative evolution operator learning, which leverages a Mamba backbone augmented with multi-patch evolutionary guidance to produce stable autonomous rollouts; and (iii) invariant measure alignment, which applies geometry-aware regularization to align generated dynamics with the system's long-term statistical properties and prevent asymptotic collapse. The details of these procedures are as follows.

\subsection{Attractor Manifold Representation} \label{sec:method-rep}
Conventional time-series forecasting models typically employ a point-wise linear projection to map observations $\bm{X}_{\text{obs}} \in \mathbb{R}^{V \times T}$ into latent embeddings $\bm{Z} \in \mathbb{R}^{V \times T \times m}$.
However, such a mapping applies a uniform linear transformation across time steps, effectively scaling the values. It fails to unfold the state space or mitigate the inherent complexity of chaotic dynamics, yielding negligible gain in representing the system's intrinsic structure.
To address this limitation, we propose a physics-informed pipeline to produce time-delay (TD) embeddings.
Grounded in Takens' embedding theorem~\cite{takens2006detecting} (see Appendix~\ref{app:represent} for background), such a method can reconstruct the attractor from a single dimension of the observation series of the underlying system~\cite{hu2024toward,wu2024predicting}. Specifically, given a chaotic system, we define two hyperparameters: the embedding dimension $m$ and the time delay $\tau$. For the univariate trajectory $\bm{x}\in \mathbb{R}^T$ of each variable, the $m$-dimensional embedding of each step $t$ is derived as:
\begin{equation} \label{equ:time_delay}
    \bm{z}_t=({x}_{t-(m-1)\tau}, {x}_{t-(m-2)\tau}, \cdots, {x}_t) \in \mathbb{R}^m,
\end{equation}
where $m$ controls the number of delayed coordinates and $\tau$ controls their temporal spacing.
In practice, $\tau$ is determined from the observed sequence by identifying the first local minimum of average mutual information~\cite{fraser1986independent}:
\begin{equation}
    \mathrm{AMI}(\tau)=
    \sum_{x_t,x_{t+\tau}}
    p(x_t,x_{t+\tau})
    \log\frac{p(x_t,x_{t+\tau})}{p(x_t)p(x_{t+\tau})},
\end{equation}
where $p(x_t)$, $p(x_{t+\tau})$, and $p(x_t,x_{t+\tau})$ denote the marginal and joint empirical distributions. This criterion encourages delayed coordinates to retain complementary dynamical information with limited redundancy.
The embedding dimension $m$ is then increased until neighborhood relations in the reconstructed manifold become stable. Following the correlation-integral criterion for nonlinear embedding-parameter selection~\cite{kim1999nonlinear}, we compute
\begin{equation}
    F(m)=\frac{1}{N-m+1}\sum_{i=1}^{N-m+1}\log C_i(m,r),
\end{equation}
where $C_i(m,r)$ counts the number of points within radius $r$ of the $i$-th point in the $m$-dimensional delay space. We choose the smallest $m$ after which $F(m)$ saturates, indicating that the delay coordinates sufficiently unfold the attractor without introducing unnecessary dimensions~\cite{kim1999nonlinear,hu2024toward}.
To ensure the $\bm{Z}$ maintains the same temporal length as $\bm{X}_{\text{obs}}$ for downstream processing, we apply zero-padding to the sequence.
This design offers two distinct theoretical advantages over standard linear embeddings.
First, TD embedding guarantees that the reconstructed geometry is diffeomorphic to the original strange attractor, in the case that $m$ is sufficiently large~\cite{takens2006detecting}, thereby rigorously recovering invariant dynamical properties.
Second, by embedding all variables, we explicitly construct distinct shadow manifolds~\cite{sugihara2012detecting}. It mitigates the fragility of single-variable embedding under finite data, noise, and partial observability.
In such non-ideal regimes, a single embedding often yields a degraded reconstruction due to variable-specific distortions or insufficient unfolding.
By integrating these views, our model leverages cross-view consistency to filter out these reconstruction artifacts, robustly capturing the true attractor features shared across all manifolds.

While the physics-informed representation $\bm{Z}$ effectively unfolds the state space, treating each time step $\bm{z}_t$ in isolation captures only static phase points on the attractor, lacking explicit information regarding the system's temporal evolution. To transition from states to continuous dynamics, we employ a patching strategy that aggregates consecutive steps.
Physically, each patch captures a short-term trajectory segment on the attractor, enabling the embedding of local dynamical patterns into the latent representation.
Formally, given the representation $\bm{Z} \in \mathbb{R}^{V\times T \times m}$, we segment it into $N=\lfloor \frac{T}{D} \rfloor$ non-overlapping patches $\bm{P} \in \mathbb{R}^{N \times (D\cdot V\cdot m)}$ with patch length $D$.
We then map each patch segment $\bm{P}_i \in \mathbb{R}^{D\cdot V\cdot m}$ to a latent embedding $\bm{S}_i \in \mathbb{R}^d$ via a linear projection:
\begin{equation}
    \bm{S}_i = \bm{W}_{\text{emb}}\bm{P}_i + \bm{b}_{\text{emb}},
\end{equation}
where $\bm{W}_{\text{emb}}$ and $\bm{b}_{\text{emb}}$ are learnable parameters. This projection serves as a learnable compressor that distills the essential geometric features of the local phase flow. Here, $\bm{P}_i$ denotes the representation patch used for tokenization, while $\bm{Y}_i=\bm{X}_{:,(i-1)D+1:iD}\in\mathbb{R}^{V\times D}$ denotes the corresponding segment in the original observed state space.

\subsection{Generative Dynamics Learning} \label{sec:method-mtp}
Reconstructing the long-term evolution of a chaotic system is fundamentally equivalent to identifying its underlying evolution operator. Unlike traditional time-series forecasting models~\cite{liu2023itransformer} that treat prediction as a discriminative sequence-to-sequence mapping, we require a generative state-space surrogate that can initialize from a short context and then evolve using its own generated states~\cite{durstewitz2023reconstructing}.
To this end, we implement the evolution operator with stacked Mamba layers, leveraging their continuous-time state-space formulation to autoregressively extrapolate latent patch tokens (see Appendix~\ref{app:ssm_mamba} for SSM and Mamba background). We employ a generative training strategy in which the model $\mathcal{M}_\theta$ evolves previous latent tokens and the decoder $\phi_{\mathrm{dec}}:\mathbb{R}^d\to\mathbb{R}^{V\times D}$ outputs the next observed-state segment:
\begin{equation}\label{equ:next-token}
    \hat{\bm{Y}}_{i+1} =
    \phi_{\mathrm{dec}}\!\left(\mathcal{M}_\theta(\bm{S}_{1}, \bm{S}_{2}, \cdots, \bm{S}_{i})\right), \quad i=1,\ldots,N-1.
\end{equation}

\noindent \textbf{Dynamical Decomposition via Residual Stacking.} Chaotic dynamics are often composed of multi-scale structures (e.g., fast-oscillating modes, slow-varying trends). To capture this inherent complexity, we implement $\mathcal{M}_\theta$ with a residual decomposition architecture~\cite{oreshkin2019n, challu2023nhits}. This design explicitly forces each Mamba layer to model specific components of the dynamics by peeling off the signal approximated by previous layers. Specifically, for the $l$-th Mamba layer, the forward pass is:
\begin{align} \mathbf{h}_{1:i}^{(l)} &= \text{Mamba}^{(l)}(\mathbf{r}_{1:i}^{(l-1)}), \quad \widehat{\bm{E}}_{1:i}^{(l)} = \text{Decoder}_e^{(l)}(\mathbf{h}_{1:i}^{(l)}),
\end{align}
where $\mathbf{r}_{1:i}^{(l)} = \mathbf{r}_{1:i}^{(l-1)} - \widehat{\bm{E}}_{1:i}^{(l)}$, and $\mathbf{r}_{1:i}^{(0)} = \bm{S}_{1:i}$ is the initial patch embedding. The final predicted observed-state segment $\widehat{\bm{Y}}_{i+1}$ is synthesized by aggregating all decomposed components:
\begin{equation}
    \widehat{\bm{Y}}_{i+1} = \phi_{\mathrm{dec}}\left(\sum_{l=1}^L \widehat{\bm{E}}_{i}^{(l)}\right).
\end{equation}
Decoders are implemented with linear layers. Such residual stacking effectively disentangles the governing laws into distinct dynamical modes, enhancing interpretability and convergence stability.

\noindent \textbf{Multi-patch Evolutionary Guidance.} A critical risk in autoregressive modeling of chaotic systems is that training only on the immediate next segment may produce hidden states that are locally accurate but insufficiently informative for autonomous long-term evolution. We therefore introduce multi-patch evolutionary guidance (MPG) as an auxiliary training constraint that asks the same evolved representation to support several future segment predictions.
Let $\mathbf{v}_i^{(0)} = \sum_{l=1}^L\hat{\bm{E}}_{i}^{(l)}$ denote the main evolved latent representation for position $i$. The main branch decodes $\mathbf{v}_i^{(0)}$ to predict the next observed-state segment $\hat{\bm{Y}}_{i+1}$. MPG then attaches $M$ auxiliary modules after this main prediction branch: the $q$-th module predicts $\hat{\bm{Y}}_{i+q+1}^{(q)}$, i.e., the $q$-th additional future segment beyond $\hat{\bm{Y}}_{i+1}$. For example, $q=1$ supervises the segment $\bm{Y}_{i+2}$.
During training, the $q$-th auxiliary module uses the previous auxiliary representation $\mathbf{v}_i^{(q-1)}$ and the teacher-forced latent embedding $\bm{S}_{i+q}$ as inputs. Its forward process is
\begin{align}
   &\widetilde{\mathbf{v}}_i^{(q)} = \psi_q([\mathrm{RMSNorm}(\mathbf{v}_i^{(q-1)}), \bm{S}_{i+q}]), \\
   &\mathbf{v}_{1:N-q-1}^{(q)} = \mathrm{Mamba}_q(\widetilde{\mathbf{v}}_{1:N-q-1}^{(q)}), \\
   &\hat{\bm{Y}}_{i+q+1}^{(q)} = \phi_{\mathrm{dec}}(\mathbf{v}_{i}^{(q)}),
\end{align}
where $N$ is the length of the input patch sequence, $\psi_q(\cdot)$ is the embedding fusion layer, and $[\cdot;\cdot]$ denotes concatenation.

\noindent \textbf{Composite Generative Training Objective.} The valid indices for the $q$-th auxiliary module are $i=1,\ldots,N-q-1$, because its target is $\bm{Y}_{i+q+1}$. The resulting next-segment and MPG losses are
\begin{align}
    \mathcal{L}_{\text{next}} &= \frac{1}{(N-1)BVD}\sum_{b=1}^B\sum_{i=1}^{N-1}\left\|\bm{Y}_{i+1}^{(b)} - \hat{\bm{Y}}_{i+1}^{(b)}\right\|_F^2, \\
    \mathcal{L}_{\text{MPG}}^q &= \frac{1}{(N-q-1)BVD}\sum_{b=1}^{B}\sum_{i=1}^{N-q-1}\left\|\bm{Y}_{i+q+1}^{(b)}-\hat{\bm{Y}}_{i+q+1}^{(q,b)}\right\|_F^2,
\end{align}
where $B$ represents the number of training samples within a batch, $q=1,\ldots,M$ with $M<N-1$ denotes the additional lookahead depth beyond the immediate next segment, and both $\mathcal{L}_{\text{next}}$ and $\mathcal{L}_{\text{MPG}}^q$ are implemented with the teacher forcing strategy~\cite{mikhaeil2022difficulty, hess2023generalized}. The composite training objective is
\begin{equation} \label{equ:gto}
    \mathcal{L} = \mathcal{L}_{\text{next}} + \frac{\lambda_p}{M}\sum_{q=1}^M\mathcal{L}_{\text{MPG}}^q,
\end{equation}
where $\lambda_p$ controls the objective importance.
We utilize multi-patch evolutionary guidance exclusively in the training process to steer the model's representation learning and enhance data efficiency. At inference, the multi-patch prediction modules are \textit{excluded}, and we rely solely on the main Mamba block for autoregressive next-patch prediction during forecasting.

\subsection{Invariant Measure Alignment}\label{sec:method-mmd}
To capture the long-term statistical behavior of chaotic systems, we introduce an invariant measure alignment stage. This stage turns the state-space model into a long-horizon generator by autoregressively producing decoded future states $\hat{\bm{X}}_{T+1:T+W}$ via student forcing. We first use MSE loss in the observed state space to anchor the trajectory to the ground truth:
\begin{equation}
    \mathcal{L}_{\text{stu}} = \frac{1}{WBV}\sum_{b=1}^B\sum_{j=T+1}^{T+W}\left\|\bm{x}_j^{(b)}-\hat{\bm{x}}_j^{(b)}\right\|_2^2.
\end{equation}
However, optimizing exclusively for point-wise accuracy in chaotic systems is prone to topological collapse, where predictions asymptotically degenerate into trivial fixed points or limit cycles. Critically, the long-horizon generation in this stage provides a sufficient collection of decoded states to approximate the system's empirical distribution in the observed phase space. This enables us to impose geometric constraints ensuring the trajectory resides on the correct attractor manifold. Specifically, we align the invariant measure, i.e., the stationary probability distribution over the phase space, of the generated trajectory with the ground truth. Let $\{\bm{u}^{(i)}\}_{i=1}^n\sim p_{\text{hist}}$, $\{\bm{v}^{(i)}\}_{i=1}^n \sim p_{\text{pred}}$, and $\{\bm{w}^{(i)}\}_{i=1}^n\sim p_{\text{gt}}$ represent observed state vectors sampled from the historical context $\bm{X}_{1:T}$, the decoded autoregressive prediction $\hat{\bm{X}}_{T+1:T+W}$, and the ground-truth future trajectory $\bm{X}_{T+1:T+W}$ within a given batch, respectively. We employ the Maximum Mean Discrepancy (MMD; see Appendix~\ref{app:mmd}) to quantify the distance between empirical distributions $p_{\text{hist}}$, $p_{\text{pred}}$, and $p_{\text{gt}}$ formed by these observed-state samples. For instance, the MMD between the future ground-truth and predicted state distributions is estimated as:
\begin{equation} \label{equ:mmd}
\widehat{\mathrm{MMD}}^2(p_{\text{gt}},p_{\text{pred}}) = \frac{1}{n^2} \left( K_{\bm{w}\bm{w}} + K_{\bm{v}\bm{v}} - 2 K_{\bm{w}\bm{v}} \right),
% \widehat{\mathrm{MMD}}^2(p_{\text{gt}},p_{\text{pred}}) = \frac{1}{n^2}(\sum_{i,j}\kappa(\bm{w^{(i)}}, \bm{w^{(j)}}) + \sum_{i,j}\kappa(\bm{v^{(i)}}, \bm{v^{(j)}}) - 2\sum_{i,j}\kappa(\bm{w^{(i)}}, \bm{v^{(j)}})).
\end{equation}
where $K_{\bm{x}\bm{y}} = \sum_{i,j}\kappa(\bm{x}^{(i)}, \bm{y}^{(j)})$.
Following prior works~\cite{schiff2024dyslim}, we implement $\kappa$ as a mixture of rational quadratic kernels. The MMD between the historical and predicted state distributions, $\widehat{\mathrm{MMD}}^2(p_{\text{hist}},p_{\text{pred}})$, is computed analogously by replacing $\{\bm{w}^{(i)}\}_{i=1}^n$ with $\{\bm{u}^{(i)}\}_{i=1}^n$. The full regularization term is:
\begin{equation}
    \mathcal{L}_{\text{reg}} = \widehat{\mathrm{MMD}}^2(p_{\text{hist}}, p_{\text{pred}}) + \lambda_{c}\widehat{\mathrm{MMD}}^2(p_{\text{gt}}, p_{\text{pred}}),
\end{equation}
where $\lambda_c$ is a hyperparameter that controls the relative importance of the two components. By simultaneously enforcing statistical stationarity relative to historical dynamics and aligning with the ground-truth distribution, they synergistically compel the predicted trajectories to strictly adhere to the intrinsic geometry of the system's attractor. The final objective for this alignment stage is:
\begin{equation}\label{equ:align}
\mathcal{L} = \mathcal{L}_{\text{stu}} + \lambda_r\mathcal{L}_{\text{reg}},
\end{equation}
where $\lambda_r$ is a hyperparameter for the regularization strength.

\subsection{Model Optimization and Inference Protocol}

\noindent \textbf{Optimization Strategy.} We devise a curriculum learning strategy to reconstruct the system's governing dynamics from local to global scales. In the initial teacher-forcing regime, the model minimizes the composite objective of next-patch prediction and temporal look-ahead guidance (Equ.~\ref{equ:gto}) to approximate the local evolution operator. Subsequently, we transition to a student-forcing regime, where the model learns to operate as an autonomous generator over long horizons. Here we minimize point-wise error while using geometry-aware MMD regularization (Equ.~\ref{equ:align}) to align the predicted dynamics with the system's invariant measure, thereby enforcing orbital stability and preventing topological collapse.

\noindent \textbf{Inference Protocol.} During inference, the optimized model functions as a surrogate for the chaotic system. Given a short context observation $\bm{X}_{\text{obs}} \in \mathbb{R}^{V\times T}$, the model encodes it into an initial sequence of latent patch tokens and initiates an autoregressive generation loop. Crucially, the auxiliary multi-patch guidance modules are excluded in this phase, allowing the core Mamba backbone to recursively evolve latent tokens based solely on the generated history. Each generated latent token is decoded into an observed-state segment, and the decoded segments are concatenated until the target horizon $H$ is reached, yielding the reconstructed future trajectory $\bm{X}_{\text{pred}} \in \mathbb{R}^{V\times H}$.

\begin{table*}[!t]
\centering
\caption{Overall mean performances on simulated and real-world datasets against trainable baselines. The best performance of each metric is marked in \textbf{bold}, and the second-best performance is \underline{underlined}. See Table~\ref{tab:zeroshot_mean} for zero-shot foundation models and Table~\ref{tab:full_overall} for 95\% CI values and complete baseline results.}
\label{tab:simulation}
\begin{adjustbox}{width=\textwidth,center}
\begin{tabular}{c|c|>{\columncolor{gray!15}}c|ccccccc|ccccc}
\toprule
\multicolumn{2}{c|}{Category}
& \multicolumn{1}{c|}{Ours}
& \multicolumn{7}{c|}{General time series forecasting model}
& \multicolumn{5}{c}{Physics-informed dynamical system model} \\
\midrule
\multicolumn{1}{c|}{System} & \multicolumn{1}{c|}{\diagbox{Metric}{Model}} & \textbf{PhyxMamba} & NBEATS & NHiTS & CrossFormer & PatchTST & TimesNet & TiDE & iTransformer & Koopa & PLRNN & nVAR & PRC & HoGRC \\
\midrule

\multirow{4}{*}{Lorenz63} 
& sMAPE ($\downarrow$) 
& \textbf{3.26} & \underline{5.34} & 13.93 & 13.62 & 64.39 & 35.84 & 84.88 & 67.84 & 60.68 & 59.26 & 94.32 & 18.34 & 18.82 \\
& $D_{\text{frac}}$ ($\downarrow$) 
& \textbf{0.060} & 0.075 & \underline{0.067} & 0.090 & 0.227 & 0.303 & 0.696 & 0.638 & 0.333 & 0.270 & 0.792 & 0.578 & 0.089 \\
& $D_{\text{stsp}}$ ($\downarrow$) 
& \textbf{1.133} & \underline{1.586} & 1.648 & 1.591 & 61.646 & 32.343 & 92.375 & 62.434 & 58.768 & 12.970 & 61.331 & 29.389 & 16.761 \\
& $D_{\text{lyap}}$ ($\downarrow$) 
& \textbf{0.198} & \underline{0.222} & 0.255 & 0.307 & 1.203 & 1.124 & 1.901 & 1.675 & 1.462 & 1.024 & 2.111 & 1.117 & 1.089 \\
\midrule

\multirow{4}{*}{Rössler} 
& sMAPE ($\downarrow$) 
& \textbf{2.84} & \underline{14.88} & 39.90 & 27.11 & 89.16 & 42.48 & 98.48 & 85.85 & 67.78 & 51.36 & 94.95 & 28.38 & 31.03 \\
& $D_{\text{frac}}$ ($\downarrow$) 
& \textbf{0.049} & \underline{0.069} & 0.088 & 0.074 & 0.158 & 0.530 & 0.455 & 0.583 & 0.497 & 0.400 & 0.750 & 0.770 & 0.611 \\
& $D_{\text{stsp}}$ ($\downarrow$) 
& \textbf{0.034} & \underline{0.137} & 0.538 & 0.187 & 8.423 & 4.985 & 15.034 & 10.807 & 5.721 & 3.950 & 7.391 & 4.549 & 4.068 \\
& $D_{\text{lyap}}$ ($\downarrow$) 
& \textbf{0.016} & \underline{0.029} & 0.030 & 0.036 & 0.122 & 0.116 & 0.169 & 0.159 & 0.101 & 0.093 & 0.128 & 0.112 & 0.114 \\
\midrule

\multirow{4}{*}{Lorenz96} 
& sMAPE ($\downarrow$) 
& \textbf{20.28} & \underline{37.51} & 51.37 & 56.64 & 116.63 & 95.93 & 123.41 & 126.08 & 114.74 & 89.90 & 86.04 & 36.42 & 47.75 \\
& $D_{\text{frac}}$ ($\downarrow$) 
& \textbf{0.164} & \underline{0.172} & 0.195 & 0.196 & 1.262 & 1.172 & 1.273 & 1.252 & 0.806 & 0.370 & 0.283 & 0.179 & 0.184 \\
& $D_{\text{stsp}}$ ($\downarrow$) 
& \textbf{15.440} & 18.043 & 18.132 & 36.191 & 141.306 & 102.414 & 174.493 & 170.792 & 128.611 & 24.020 & 23.537 & \underline{16.834} & 17.287 \\
& $D_{\text{lyap}}$ ($\downarrow$) 
& \textbf{0.204} & 0.233 & 0.222 & 0.493 & 1.691 & 1.386 & 1.820 & 1.147 & 1.587 & 0.394 & 2.225 & \underline{0.216} & 0.219 \\
\midrule

\multirow{4}{*}{ECG} 
& sMAPE ($\downarrow$) 
& \textbf{37.14} & 50.77 & \underline{50.64} & 58.70 & 64.83 & 60.84 & 136.53 & 55.59 & 60.63 & 116.00 & 147.87 & 136.60 & 144.10 \\
& $D_{\text{frac}}$ ($\downarrow$) 
& \textbf{0.036} & \underline{0.038} & 0.078 & 0.443 & 0.218 & 0.143 & 0.219 & 0.242 & 0.245 & 0.529 & 0.642 & 0.147 & 0.184 \\
& $D_{\text{stsp}}$ ($\downarrow$) 
& \textbf{0.004} & \underline{0.034} & 0.035 & 0.156 & 0.440 & 0.682 & 0.834 & 0.439 & 0.463 & 1.592 & 0.638 & 3.265 & 4.015 \\
& $D_{\text{lyap}}$ ($\downarrow$) 
& \textbf{0.230} & 0.453 & 0.421 & \underline{0.373} & 1.501 & 2.212 & 1.800 & 1.549 & 1.244 & 6.109 & 6.858 & 7.565 & 7.632 \\
\midrule

\multirow{4}{*}{Worm} 
& sMAPE ($\downarrow$) 
& \textbf{89.51} & 113.89 & 116.05 & \underline{90.18} & 113.52 & 100.34 & 120.14 & 134.22 & 110.70 & 122.01 & 111.84 & 102.45 & 105.12 \\
& $D_{\text{frac}}$ ($\downarrow$) 
& \textbf{0.159} & 0.194 & \underline{0.184} & 0.294 & 0.974 & 0.815 & 0.915 & 0.806 & 0.909 & 0.638 & 1.051 & 0.855 & 0.890 \\
& $D_{\text{stsp}}$ ($\downarrow$) 
& \textbf{9.875} & 15.495 & \underline{11.671} & 19.332 & 41.069 & 26.837 & 44.876 & 34.794 & 36.829 & 25.474 & 24.410 & 18.550 & 20.125 \\
& $D_{\text{lyap}}$ ($\downarrow$) 
& \textbf{0.079} & 0.104 & \underline{0.097} & 0.134 & 0.672 & 0.450 & 0.621 & 0.604 & 0.602 & 0.313 & 0.721 & 0.510 & 0.545 \\
\midrule

\multirow{4}{*}{ENSO} 
& sMAPE ($\downarrow$) 
& \textbf{40.21} & 47.02 & 72.18 & 64.78 & 51.22 & 48.25 & 53.36 & 49.63 & \underline{44.45} & 90.01 & 52.16 & 45.33 & 48.90 \\
& $D_{\text{frac}}$ ($\downarrow$) 
& \textbf{0.230} & 0.717 & 0.734 & \underline{0.265} & 0.377 & 0.384 & 0.499 & 0.324 & 0.334 & 0.986 & 0.320 & 0.285 & 0.299 \\
& $D_{\text{stsp}}$ ($\downarrow$) 
& \textbf{0.451} & 7.613 & 31.120 & 42.974 & 6.478 & 6.632 & 6.366 & 6.483 & 5.600 & 5.416 & 1.264 & \underline{1.150} & 1.210 \\
& $D_{\text{lyap}}$ ($\downarrow$) 
& \textbf{0.006} & 0.013 & 0.008 & 0.013 & 0.013 & \underline{0.007} & 0.015 & 0.010 & 0.009 & 0.031 & 0.012 & 0.012 & 0.011 \\
\midrule

\bottomrule
\end{tabular}
\end{adjustbox}

\end{table*}

\begin{table}[!t]
\centering
\caption{Mean performance comparison with zero-shot foundation models. The best performance of each metric is marked in \textbf{bold}. See Table~\ref{tab:full_overall} for 95\% CI values and complete baseline results.}
\label{tab:zeroshot_mean}
\begin{adjustbox}{width=\columnwidth,center}
\begin{tabular}{c|c|>{\columncolor{gray!15}}c|ccccc}
\toprule
\multicolumn{2}{c|}{Category}
& \multicolumn{1}{c|}{Ours}
& \multicolumn{5}{c}{Zero-shot foundation model} \\
\midrule
System & \diagbox{Metric}{Model} & \textbf{PhyxMamba} & Panda & Chronos2 & Chronos2-S & DynaMix & Parrot \\
\midrule

\multirow{4}{*}{Lorenz63}
& sMAPE ($\downarrow$)
& \textbf{3.26} & 105.71 & 92.18 & 93.26 & 109.60 & 112.61 \\
& $D_{\text{frac}}$ ($\downarrow$)
& \textbf{0.060} & 0.157 & 0.123 & 0.147 & 0.326 & 0.136 \\
& $D_{\text{stsp}}$ ($\downarrow$)
& \textbf{1.133} & 8.229 & 8.956 & 9.196 & 10.110 & 5.150 \\
& $D_{\text{lyap}}$ ($\downarrow$)
& \textbf{0.198} & 0.937 & 0.785 & 0.827 & 0.540 & 0.556 \\
\midrule

\multirow{4}{*}{Rössler}
& sMAPE ($\downarrow$)
& \textbf{2.84} & 154.48 & 87.56 & 85.74 & 160.56 & 94.59 \\
& $D_{\text{frac}}$ ($\downarrow$)
& \textbf{0.049} & 0.141 & 0.082 & 0.069 & 0.370 & 0.093 \\
& $D_{\text{stsp}}$ ($\downarrow$)
& \textbf{0.034} & 4.611 & 3.036 & 3.541 & 4.952 & 0.553 \\
& $D_{\text{lyap}}$ ($\downarrow$)
& \textbf{0.016} & 0.062 & 0.040 & 0.039 & 0.106 & 0.039 \\
\midrule

\multirow{4}{*}{Lorenz96}
& sMAPE ($\downarrow$)
& \textbf{20.28} & 138.20 & 117.57 & 122.05 & 139.52 & 135.25 \\
& $D_{\text{frac}}$ ($\downarrow$)
& \textbf{0.164} & 0.669 & 0.392 & 0.179 & 0.716 & 0.880 \\
& $D_{\text{stsp}}$ ($\downarrow$)
& \textbf{15.440} & 20.227 & 20.269 & 20.228 & 20.479 & 20.450 \\
& $D_{\text{lyap}}$ ($\downarrow$)
& \textbf{0.204} & 0.543 & 0.249 & 0.299 & 0.390 & 0.404 \\
\midrule

\multirow{4}{*}{ECG}
& sMAPE ($\downarrow$)
& \textbf{37.14} & 150.39 & 150.21 & 156.02 & 163.27 & 61.30 \\
& $D_{\text{frac}}$ ($\downarrow$)
& \textbf{0.036} & 0.328 & 0.407 & 0.642 & 1.018 & 0.099 \\
& $D_{\text{stsp}}$ ($\downarrow$)
& \textbf{0.004} & 0.566 & 0.775 & 0.715 & 1.853 & 0.144 \\
& $D_{\text{lyap}}$ ($\downarrow$)
& \textbf{0.230} & 0.464 & 0.510 & 0.481 & 1.020 & 0.367 \\
\midrule

\multirow{4}{*}{Worm}
& sMAPE ($\downarrow$)
& \textbf{89.51} & 144.16 & 116.59 & 114.07 & 144.74 & 123.11 \\
& $D_{\text{frac}}$ ($\downarrow$)
& \textbf{0.159} & 0.622 & 0.443 & 0.401 & 1.069 & 1.230 \\
& $D_{\text{stsp}}$ ($\downarrow$)
& \textbf{9.875} & 12.321 & 12.352 & 12.011 & 11.621 & 10.678 \\
& $D_{\text{lyap}}$ ($\downarrow$)
& \textbf{0.079} & 0.111 & 0.086 & 0.082 & 0.174 & 0.209 \\
\midrule

\multirow{4}{*}{ENSO}
& sMAPE ($\downarrow$)
& \textbf{40.21} & 57.70 & 49.74 & 48.89 & 47.67 & 44.70 \\
& $D_{\text{frac}}$ ($\downarrow$)
& \textbf{0.230} & 0.532 & 0.476 & 0.524 & 0.374 & 0.455 \\
& $D_{\text{stsp}}$ ($\downarrow$)
& \textbf{0.451} & 4.891 & 4.371 & 4.583 & 6.481 & 6.001 \\
& $D_{\text{lyap}}$ ($\downarrow$)
& \textbf{0.006} & 0.012 & 0.011 & 0.010 & 0.013 & 0.015 \\
\midrule

\bottomrule
\end{tabular}
\end{adjustbox}
\end{table}

\section{Experiments} \label{sec:exp}
\subsection{Experimental Setup} \label{sec:exp-setup}
\textbf{Datasets.} We evaluate our method on both simulated and real-world chaotic systems. The simulated benchmark consists of three canonical ordinary differential equation systems with representative strange attractors and different levels of dynamical complexity~\cite{mikhaeil2022difficulty, hess2023generalized}.

\begin{itemize}[leftmargin=*,itemsep=0.2em,topsep=0.2em,parsep=0pt,partopsep=0pt]
\item \textbf{Lorenz63 system.} Lorenz63~\cite{lorenz2017deterministic} is a prototypical low-dimensional chaotic flow that exhibits the well-known butterfly-shaped strange attractor. It is defined as:
\begin{equation*}
\begin{aligned}
    \dot{x} &= \sigma(y-x),\\
    \dot{y} &= \rho x-y-xz,\\
    \dot{z} &= xy-\beta z.
\end{aligned}
\end{equation*}
We use the standard chaotic parameter setting $\sigma=10$, $\rho=28$, and $\beta=8/3$.

\item \textbf{Rössler system.} The Rössler system~\cite{rossler1976equation} is another classical continuous-time chaotic system whose attractor has a spiral structure with a relatively simple phase-space geometry. It is governed by:
\begin{equation*}
\begin{aligned}
    \dot{x} &= -(y+z),\\
    \dot{y} &= x+\alpha y,\\
    \dot{z} &= \beta+z(x-\gamma).
\end{aligned}
\end{equation*}
Following the standard chaotic regime, we set $\alpha=0.2$, $\beta=0.2$, and $\gamma=5.7$.

\item \textbf{Lorenz96 system.} Lorenz96~\cite{lorenz1996predictability} is a cyclically coupled dynamical system widely used to study chaotic evolution and predictability. Its dynamics are:
\begin{equation*}
\begin{aligned}
    \dot{x}_k &= (x_{k+1}-x_{k-2})x_{k-1}\\
    &\quad -x_k+F,\quad k=1,\ldots,N,
\end{aligned}
\end{equation*}
where the variables follow cyclic boundary conditions. We take $N=5$ and $F=20$, locating the system in the chaotic regime.
\end{itemize}

Trajectories are produced with a fourth-order Runge-Kutta integrator using $\Delta t=0.001$ based on the \texttt{dyst} library~\cite{gilpin2021chaos}, and are then downsampled to a unified temporal granularity of approximately 30 points per Lyapunov time. For each simulated system, we generate long trajectories of 30,000 time points (about 1,000 Lyapunov times) for training/validation segment construction and use 100 independently sampled initial conditions for testing. We further conduct experiments on the Forced FitzHugh-Nagumo and Hindmarsh-Rose systems; their system definitions and complete results are provided in Appendix~\ref{app:dataset} and Appendix~\ref{sec:addition_synthetic}, respectively.

% We also introduce two PDE-based systems, Kuramoto–Sivashinsky (KS) and Von Kármán Vortex Street (VKVS) dynamics, as representative cases for high-dimensional chaos. 
\noindent For the real-world chaotic datasets, we further evaluate our method on three empirical datasets that exhibit complex nonlinear dynamics across physiological, behavioral, and climate systems, demonstrated as follows.
\begin{itemize}[leftmargin=*,itemsep=0.2em,topsep=0.2em,parsep=0pt,partopsep=0pt]
\item \textbf{Electrocardiogram (ECG).} The ECG dataset~\cite{reiss2019deep} records scalar physiological signals corresponding to heart muscle potentials. Following prior work, we embed the scalar time series into a 5-dimensional state space using the PECUZAL algorithm~\cite{kramer2021unified}, enabling attractor-based reconstruction from physiological dynamics; its positive maximum Lyapunov exponent reported in prior work supports chaotic dynamics~\cite{mikhaeil2022difficulty}.

\item \textbf{Worm motion.} The Worm dataset consists of locomotion recordings from \textit{Caenorhabditis elegans} in a foraging setting. The original high-dimensional body-shape representation is reduced to the five dominant eigenworm components; prior analysis reports a positive maximum Lyapunov exponent, indicating chaotic behavior~\cite{ahamed2021capturing}.

\item \textbf{ENSO.} The ENSO dataset~\cite{wu2024predicting} provides weekly oceanic indicators for the El Niño-Southern Oscillation from September 1981 to October 2022. For each of four equatorial Pacific regions, it contains two variables: sea surface temperature (SST), which measures the absolute ocean-surface temperature, and sea surface temperature anomaly (SSTA), which measures the deviation from the long-term climatological mean. The ENSO ocean-atmosphere oscillator has been characterized as chaotic~\cite{tziperman1994el}, making it a challenging benchmark.
\end{itemize}
For train/test splits, ECG uses one trajectory for training/validation and the other for testing; Worm uses recordings from eight worms for training and the remaining two worms for testing; ENSO adopts a chronological split, using the first 80\% of the time series for training and the remaining 20\% for testing. Further dataset-specific preprocessing and split details are provided in Appendix~\ref{app:dataset}.
These datasets cover controlled ODE systems, biological rhythms, animal locomotion, and climate dynamics, allowing us to evaluate whether a model learned from short observations can reconstruct long-term attractor behavior across heterogeneous sources.

\noindent\textbf{Baselines.} 
Baseline models can be categorized into three primary paradigms:
\begin{itemize}[leftmargin=*,itemsep=0.2em,topsep=0.2em,parsep=0pt,partopsep=0pt]
\item \textbf{General time series forecasting model.} This category includes NBEATS~\cite{oreshkin2019n}, NHiTS~\cite{challu2023nhits}, CrossFormer~\cite{zhang2023crossformer}, PatchTST~\cite{nie2022time}, TimesNet~\cite{wu2022timesnet}, TiDE~\cite{das2023long}, and iTransformer~\cite{liu2023itransformer}. They cover representative MLP-, CNN-, and Transformer-style forecasting architectures, allowing us to test whether strong short-horizon predictors can preserve long-term chaotic geometry.

\item \textbf{Physics-informed dynamical system model.} This category includes Koopa~\cite{liu2023koopa}, PLRNN~\cite{mikhaeil2022difficulty}, nVAR~\cite{gauthier2021next}, PRC~\cite{srinivasan2022parallel}, and HoGRC~\cite{li2024higher}. These methods are designed to model nonlinear state evolution, Koopman-style latent dynamics, or reservoir-style dynamics, and thus serve as dedicated baselines for chaotic system reconstruction.

\item \textbf{Zero-shot foundation model.} This category includes Panda~\cite{lai2025panda}, Chronos2 and its smaller counterpart (Chronos2-S)~\cite{ansari2025chronos}, DynaMix~\cite{hemmer2025true}, and Parrot~\cite{zhang2025context}. These models evaluate whether generic time-series priors learned from large-scale pretraining can reconstruct chaotic dynamics from short context observations without task-specific training.
\end{itemize}
All trainable baselines are evaluated under the same short-context reconstruction setting, and the comparison reports both point-wise accuracy and attractor-level fidelity.

\noindent\textbf{Experimental Settings.} 
Our goal is to infer an underlying dynamical generator from short context observations and to faithfully reproduce the system's long-term behavior under autonomous rollouts. To avoid conflating short context with short total training data, we train models on short segments sampled from longer trajectories. Observational data for ODE-based chaotic systems and real-world datasets are resampled by their Lyapunov time ($T_L$), with each Lyapunov time corresponding to 30 time steps.
During teacher-forcing training, each sample spans one Lyapunov time ($1T_L$) and is used for next-patch prediction, where the patch length $D < T_L$. During student forcing, each training segment spans $2T_L$: the first $T_L$ provides the observation context and the following $T_L$ provides the autoregressive reconstruction target. Baseline models are trained under the same short-context protocol. For evaluation, we provide each model with an observation context of length $T_L$ and task it with autonomously rolling out the subsequent trajectory of length $10T_L$. 
% For the PDE-based VKVS dynamics, we adopt a fixed window of 128 steps rather than rescaling with Lyapunov time, which corresponds to multiple vortex shedding cycles and provides a physically meaningful temporal context. The model is trained to predict its next 128 steps during student forcing and is further tasked with generating a long-term trajectory of 1,280 steps during evaluation.

\noindent\textbf{Evaluation Metrics.} 
We employ a suite of evaluation metrics that capture both short-term predictive accuracy and long-term statistical fidelity of the reconstructed chaotic systems.
Following the notation in Section~\ref{sec:pfp}, for the $s$-th test sample, let $\bm{X}_{T+1:T+H}^{(s)}=\{\bm{x}^{(s)}_{T+h}\}_{h=1}^{H}$ and $\hat{\bm{X}}_{T+1:T+H}^{(s)}=\{\hat{\bm{x}}^{(s)}_{T+h}\}_{h=1}^{H}$ denote the ground-truth future trajectory and the model-reconstructed autonomous rollout, respectively. Here, $T$ is the context length, $H$ is the rollout horizon, and $S$ is the number of evaluated test samples.
\begin{itemize}[leftmargin=*,itemsep=0.2em,topsep=0.2em,parsep=0pt,partopsep=0pt]
\item \textbf{Symmetric absolute percentage error (sMAPE).} We compute sMAPE over the first 10\% of the predicted trajectory (1$T_L$), corresponding to the short-term predictable regime. Let $H_{\mathrm{short}}=\lfloor0.1H\rfloor$ denote this short-term horizon:
\begin{equation*}
    \mathrm{sMAPE}=2\frac{100}{H_{\mathrm{short}}}
    \sum_{h=1}^{H_{\mathrm{short}}}
    \frac{|\bm{x}_{T+h}-\hat{\bm{x}}_{T+h}|}
    {|\bm{x}_{T+h}| + |\hat{\bm{x}}_{T+h}|},
\end{equation*}
where $\bm{x}_{T+h}$ and $\hat{\bm{x}}_{T+h}$ denote the ground-truth and predicted future states.

\item \textbf{Correlation dimension error ($D_{\text{frac}}$).} We compute $D_{\text{frac}}$ over the entire autonomous rollout (10$T_L$). It measures the discrepancy between the fractal dimensions of the true and reconstructed attractors, reflecting whether the learned dynamics reproduce the intrinsic geometric complexity of the system. For any rollout $\bm{X}_{T+1:T+H}=\{\bm{x}_{T+h}\}_{h=1}^{H}$, we first compute the correlation sum
\begin{equation*}
    C_{\bm{X}}(r)=\frac{2}{H(H-1)}
    \sum_{1\le i<j\le H}\mathbb{I}\!\left(\|\bm{x}_{T+i}-\bm{x}_{T+j}\|_2<r\right),
\end{equation*}
and estimate its correlation dimension $d_c(\bm{X})$ as the slope of $\log C_{\bm{X}}(r)$ with respect to $\log r$ over the scaling region:
\begin{equation*}
    d_c(\bm{X})=\frac{d\log C_{\bm{X}}(r)}{d\log r}.
\end{equation*}
The resulting error is defined as
\begin{equation*}
\begin{aligned}
    D_{\text{frac}} =
    \sqrt{\frac{1}{S}\sum_{s=1}^{S}
    \left(d_c(\bm{X}_{T+1:T+H}^{(s)})-d_c(\hat{\bm{X}}_{T+1:T+H}^{(s)})\right)^2},
\end{aligned}
\end{equation*}

\item \textbf{Attractor KL divergence ($D_{\text{stsp}}$).} We compute $D_{\text{stsp}}$ over the entire autonomous rollout (10$T_L$). It compares invariant phase-space distributions by fitting Gaussian mixture models to the generated and ground-truth trajectories and estimating their KL divergence. Following~\cite{zhang2024zero}, for any rollout $\bm{X}_{T+1:T+H}=\{\bm{x}_{T+h}\}_{h=1}^{H}$, we construct its empirical phase-space density by centering one Gaussian component at each evaluated future state:
\begin{equation*}
    q_{\bm{X}}(\bm{u}) =
    \frac{1}{H}\sum_{h=1}^{H}
    \mathcal{N}(\bm{u};\bm{x}_{T+h},\bm{\Sigma}_{T+h}),
    \quad \bm{u}\in\mathbb{R}^{V}.
\end{equation*}
Thus, the number of mixture components is the number of evaluated future states within a single rollout, i.e., $H$, rather than the number of test samples or a separately tuned model hyperparameter. The same covariance-setting protocol is applied to all ground-truth and reconstructed rollouts. For the $s$-th test sample, we set $q_{\text{gt}}^{(s)}=q_{\bm{X}_{T+1:T+H}^{(s)}}$ and $q_{\text{pred}}^{(s)}=q_{\hat{\bm{X}}_{T+1:T+H}^{(s)}}$. The attractor-level divergence is then defined as
\begin{equation*}
\begin{aligned}
    D_{\text{stsp}} =
    \frac{1}{S}\sum_{s=1}^{S}
    D_{\mathrm{KL}}\!\left(q_{\text{gt}}^{(s)}\,\|\,q_{\text{pred}}^{(s)}\right),
\end{aligned}
\end{equation*}

\item \textbf{Maximum Lyapunov exponent error ($D_{\text{lyap}}$).} We compute $D_{\text{lyap}}$ over the entire autonomous rollout (10$T_L$). It compares the sensitivity to initial conditions of the reconstructed and true systems. Let $\lambda_{\max,\mathrm{gt}}^{(s)}$ and $\lambda_{\max,\mathrm{pred}}^{(s)}$ denote the maximum Lyapunov exponents estimated from $\bm{X}_{T+1:T+H}^{(s)}$ and $\hat{\bm{X}}_{T+1:T+H}^{(s)}$, respectively, using the Rosenstein algorithm~\cite{rosenstein1993practical}. We define
\begin{equation*}
    D_{\text{lyap}} =
    \sqrt{\frac{1}{S}\sum_{s=1}^{S}
    \left(\lambda_{\max,\mathrm{gt}}^{(s)}
    -\lambda_{\max,\mathrm{pred}}^{(s)}\right)^2}.
\end{equation*}
\end{itemize}
Lower values indicate better reconstruction fidelity for all metrics.

\begin{figure}[!t]
    \centering
    \includegraphics[width=\columnwidth]{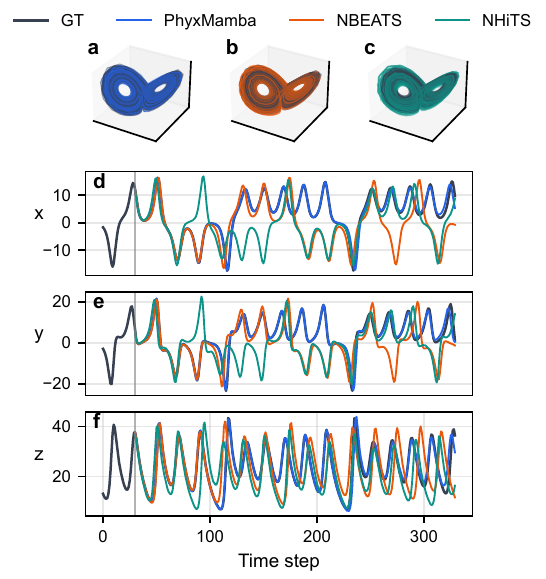}
    \caption{Attractor and representative trajectories for the Lorenz63 system, comparing GT with PhyxMamba, NBEATS, and NHiTS. The vertical line marks the start of prediction.}
    \label{fig:visualization}
\end{figure}

\subsection{Overall Performance}

Table~\ref{tab:simulation} reports mean performance against trainable baselines, including general time series forecasting models and physics-informed dynamical system models. Table~\ref{tab:zeroshot_mean} reports mean performance against zero-shot foundation models. The complete results with all baselines and 95\% confidence intervals are provided in Table~\ref{tab:full_overall} of Appendix~\ref{app:overall}.
Figure~\ref{fig:visualization} provides a qualitative visualization of a representative reconstruction case of the Lorenz63 system. Our key findings are as follows:

\begin{itemize}[leftmargin=*,itemsep=0.2em,topsep=0.2em,parsep=0pt,partopsep=0pt]
\item \textbf{State-of-the-art Short-term Forecasting Accuracy.} Our model consistently achieves superior short-term forecasting performance, recording the lowest sMAPE across all six evaluated datasets. PhyxMamba significantly outperforms both classical recurrent models and advanced Transformers. For instance, on the Lorenz63 system, PhyxMamba achieves an sMAPE of \textbf{3.26}, reducing the error by approximately 39\% compared to the second-best NBEATS (\underline{5.34}). Similarly, in the Lorenz96 system, our model maintains this advantage with an sMAPE of \textbf{20.28}, substantially lower than \underline{36.42} achieved by PRC. These results indicate that PhyxMamba effectively captures the instantaneous trajectory evolution of chaotic systems.

\item \textbf{Preservation of Long-term Statistical and Geometric Properties.} Beyond the predictable horizon, PhyxMamba excels as a surrogate model, preserving the invariant properties and geometric integrity of the underlying system. This is evidenced by its performance on $D_{\text{frac}}$, $D_{\text{stsp}}$, and $D_{\text{lyap}}$ metrics, where it achieves the best results. On the Rössler system, PhyxMamba yields a Lyapunov exponent error ($D_{\text{lyap}}$) of \textbf{0.016} and a correlation dimension error ($D_{\text{frac}}$) of \textbf{0.049}, surpassing the second-best baselines (\underline{0.029} and \underline{0.069}, respectively). Furthermore, on the challenging real-world ENSO dataset, our model achieves a KL divergence ($D_{\text{stsp}}$) of \textbf{0.451}, an order of magnitude lower than the runner-up PRC (\underline{1.150}). These results confirm that PhyxMamba does not merely memorize trajectories but reconstructs the global phase-space structure and the chaotic nature of the target systems.

\item \textbf{Disconnection Between Prediction Error and Dynamic Fidelity.}
A critical insight from our experiments corroborates the findings of Durstewitz et al.~\cite{durstewitz2023reconstructing}: a comparatively low prediction error effectively measures curve-fitting capability but does not guarantee agreement between the true and reconstructed dynamical systems.
It is clearly observable in the ENSO dataset results. While Koopa achieves a highly competitive sMAPE of \underline{44.45} (close to PhyxMamba's \textbf{40.21}), it fails remarkably in capturing the system's topology, exhibiting a $D_{\text{stsp}}$ of 5.600—more than $12\times$ worse than PhyxMamba (\textbf{0.451}). Similarly, on the Worm dataset, CrossFormer achieves the second-best sMAPE (\underline{90.18}) but incurs a $D_{\text{stsp}}$ of 19.332, nearly double the error of our model (\textbf{9.875}).
As noted in~\cite{durstewitz2023reconstructing}, standard deep learning models often overfit to short-term temporal correlations while violating the fundamental geometric and ergodic properties of the system. In contrast, PhyxMamba harmonizes both objectives, ensuring accurate forecasts that are consistent with ground-truth dynamics.

\item \textbf{Short Context Limits Zero-shot System Reconstruction.} Table~\ref{tab:zeroshot_mean} shows that zero-shot foundation models degrade within the short context reconstruction protocol: on Lorenz63, the best zero-shot sMAPE is 92.18, far above PhyxMamba's \textbf{3.26}; on ENSO, the best zero-shot $D_{\text{stsp}}$ is 4.371, whereas PhyxMamba attains \textbf{0.451}. These results show that under short context observation settings, zero-shot in-context adaptation is insufficient for recovering system-specific dynamics.
\end{itemize}

\subsection{Ablation Study}
\subsubsection{\textbf{Pipeline Component Ablation}}
\begin{figure}[t]
    \centering
    \includegraphics[width=\columnwidth]{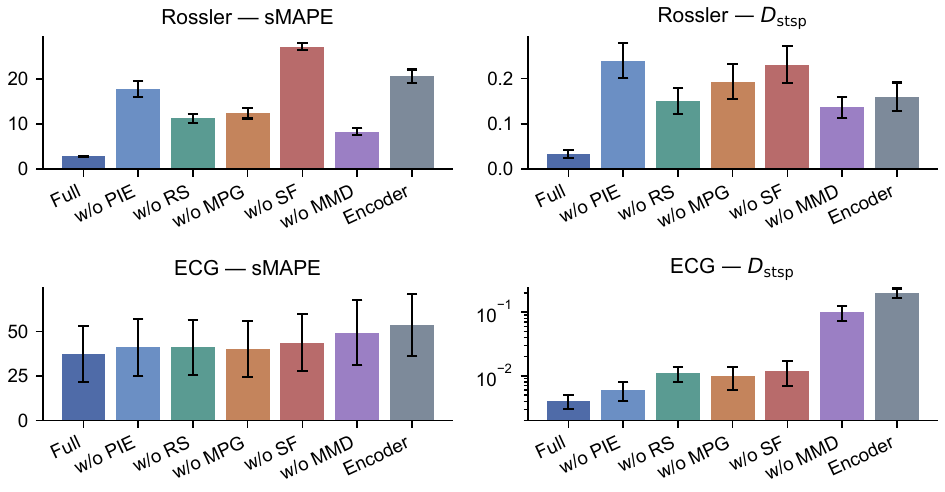}
    \caption{Ablation study on Rössler and ECG. Bars show mean ± 95\% CI over runs with different initial conditions.}
    \label{fig:ablation}
\end{figure}
To evaluate the contribution of each design element, all ablation variants are trained and evaluated under the same short-context reconstruction protocol as the full model; each variant changes one component while keeping the remaining pipeline unchanged. We consider six variants:
\begin{itemize}[leftmargin=*,itemsep=0.15em,topsep=0.15em,parsep=0pt,partopsep=0pt]
\item \textbf{Encoder-only Mamba (Encoder).} To isolate the role of the generative autoregressive formulation, we replace it with an encoder-only Mamba. Historical observations are encoded by Mamba blocks, flattened, and projected by linear layers for direct sequence-level forecasting; inference recursively feeds each predicted segment back as the next context.
\item \textbf{Without physics-informed embedding (w/o PIE).} The physics-informed time-delay embedding module is bypassed, and patching and embedding are applied directly to the observed sequence.
\item \textbf{Without residual stacking (w/o RS).} In this setting, the residual-stacked hierarchical Mamba block is replaced with a plain layer-by-layer stacked Mamba block, testing the value of decomposed forecasting.
\item \textbf{Without multi-patch evolutionary guidance (w/o MPG).} Multi-patch evolutionary guidance is removed, so the backbone is trained only with next-patch prediction.
\item \textbf{Without student forcing (w/o SF).} We disable the student-forcing stage and retain only teacher-forcing training.
\item \textbf{Without maximum mean discrepancy regularization (w/o MMD).} Maximum mean discrepancy (MMD) regularization is excluded from the student-forcing stage to assess the effect of invariant measure alignment.
\end{itemize}
This setting separately examines the effects of phase-space unfolding, hierarchical residual state evolution, multi-step evolutionary supervision, rollout-aware training, invariant-measure alignment, and the generative reconstruction formulation. As evidenced in Figure~\ref{fig:ablation}, for the Rössler system, w/o PIE and w/o SF result in the most substantial degradation of attractor geometric integrity ($D_{\text{stsp}}$) and point-wise accuracy (sMAPE), respectively, validating the necessity of phase-space unfolding and error accumulation mitigation. Invariant measure alignment is also critical for avoiding collapse in long-term distributional fidelity. Combined with the complete numerical results in Table~\ref{tab:ablation_r_ecg} of Appendix~\ref{app:ablation_results}, PhyxMamba’s holistic integration of geometric priors, evolutionary stability, and statistical alignment is essential for robust reconstruction across varying chaotic systems. Notably, the consistent failure of Encoder-only Mamba validates that a faithful surrogate model must generatively reproduce the system's underlying evolutionary operator; superficial sequence-to-sequence correlations are insufficient for chaotic system reconstruction.

\subsubsection{\textbf{Backbone Ablation}}
\begin{table}[!t]
\centering
\caption{Controlled ablation of backbone and pipeline design on Lorenz96. Reported values are mean $\pm$ 95\% CI.}
\label{tab:backbone_pipeline_ablation}
\setlength{\tabcolsep}{2.5pt}
\resizebox{\columnwidth}{!}{%
\begin{tabular}{cccccc}
\toprule
Backbone & Pipeline design & sMAPE ($\downarrow$) & $D_{\text{frac}}$ ($\downarrow$) & $D_{\text{stsp}}$ ($\downarrow$) & $D_{\text{lyap}}$ ($\downarrow$) \\
\midrule
\rowcolor{gray!15}
\textbf{Mamba} & \textbf{Full}
& \textbf{20.28 $\pm$ 2.69}
& \textbf{0.164 $\pm$ 0.020}
& \textbf{15.440 $\pm$ 0.799}
& \textbf{0.204 $\pm$ 0.028} \\
Mamba & None
& \underline{25.97 $\pm$ 2.84}
& \underline{0.197 $\pm$ 0.020}
& \underline{17.045 $\pm$ 0.781}
& \underline{0.230 $\pm$ 0.031} \\
Transformer & Full
& 31.26 $\pm$ 3.31
& 0.218 $\pm$ 0.024
& 18.747 $\pm$ 0.826
& 0.235 $\pm$ 0.030 \\
GRU & Full
& 26.70 $\pm$ 3.22
& 0.212 $\pm$ 0.022
& 17.939 $\pm$ 0.737
& 0.242 $\pm$ 0.033 \\
TCN & Full
& 33.15 $\pm$ 3.55
& 0.204 $\pm$ 0.020
& 19.104 $\pm$ 0.823
& 0.240 $\pm$ 0.033 \\
\bottomrule
\end{tabular}}
\end{table}

To further disentangle the contribution of the state-space backbone from the pipeline design, Table~\ref{tab:backbone_pipeline_ablation} reports controlled comparisons on Lorenz96. In the Pipeline design column, Full denotes the complete PhyxMamba pipeline, where time-delay embedding, multi-patch evolutionary guidance, student forcing, and MMD regularization are enabled around the selected backbone. None denotes a plain forecasting-style generative backbone: the raw short-context observations are first mapped into the model hidden dimension by a standard MLP input projection, processed by the backbone, and then decoded back to the observed variables through an output projection, following common practice in time-series forecasting. In this comparison, the time-delay phase-space construction and the rollout-level pipeline modules are disabled. Thus, the Mamba--Full row corresponds to PhyxMamba, namely our full design. PhyxMamba achieves the best performance across all metrics, reducing sMAPE from 25.97 for plain generative Mamba to \textbf{20.28} and lowering $D_{\text{stsp}}$ from 17.045 to \textbf{15.440}, confirming that the proposed pipeline design provides essential gains beyond the backbone. Replacing Mamba with Transformer, GRU~\cite{cho2014learning}, or TCN~\cite{bai2018empirical} while keeping the same Full pipeline design consistently degrades both forecasting accuracy and dynamical fidelity; for example, Transformer--Full increases sMAPE to 31.26 and $D_{\text{stsp}}$ to 18.747. Meanwhile, plain generative Mamba remains stronger than the alternative backbones with the Full pipeline design, indicating that Mamba supplies a favorable dynamical modeling inductive bias for system evolution. These controlled comparisons show that PhyxMamba's improvement comes from the coupling between a suitable state-space backbone and the proposed pipeline design.

\begin{table}[!t]
\centering
\caption{Performance comparison between competitive baselines under partial observations for the Lorenz63 system.}
\label{tab:partial_obs_smape_kld}
\setlength{\tabcolsep}{3pt}
\resizebox{\columnwidth}{!}{%
\begin{tabular}{c|>{\columncolor{gray!15}}c>{\columncolor{gray!15}}c|cc|cc}
\toprule
\multirow{2}{*}{Observation} 
& \multicolumn{2}{>{\columncolor{gray!15}}c|}{\textbf{PhyxMamba}}
& \multicolumn{2}{c|}{NBEATS} 
& \multicolumn{2}{c}{NHiTS} \\
\cmidrule(lr){2-3} \cmidrule(lr){4-5} \cmidrule(lr){6-7}
& sMAPE ($\downarrow$) & $D_{\text{stsp}}$ ($\downarrow$)
& sMAPE ($\downarrow$) & $D_{\text{stsp}}$ ($\downarrow$)
& sMAPE ($\downarrow$) & $D_{\text{stsp}}$ ($\downarrow$) \\
\midrule

$x$   
& \textbf{10.16 $\pm$ 2.81} & \textbf{0.093 $\pm$ 0.024}
& 81.88 $\pm$ 9.76 & 10.437 $\pm$ 2.902
& \underline{31.78 $\pm$ 5.72} & \underline{0.133 $\pm$ 0.025} \\

$y$   
& \textbf{15.64 $\pm$ 2.91} & \textbf{0.097 $\pm$ 0.022}
& \underline{25.72 $\pm$ 5.01} & \underline{0.108 $\pm$ 0.021}
& 42.68 $\pm$ 5.55 & 0.909 $\pm$ 0.842 \\

$z$   
& \textbf{1.91 $\pm$ 0.42} & \textbf{0.070 $\pm$ 0.017}
& 10.55 $\pm$ 2.19 & 0.784 $\pm$ 0.128
& \underline{6.45 $\pm$ 1.31} & \underline{0.097 $\pm$ 0.069} \\

$x,y$ 
& \textbf{11.65 $\pm$ 2.79} & \textbf{0.312 $\pm$ 0.072}
& \underline{14.45 $\pm$ 3.78} & \underline{0.699 $\pm$ 0.613}
& 32.34 $\pm$ 5.91 & 1.243 $\pm$ 0.929 \\

$x,z$ 
& \textbf{5.02 $\pm$ 0.99} & \textbf{0.619 $\pm$ 0.136}
& \underline{11.13 $\pm$ 2.18} & \underline{1.566 $\pm$ 1.015}
& 41.08 $\pm$ 5.24 & 2.043 $\pm$ 0.277 \\

$y,z$ 
& \textbf{6.72 $\pm$ 1.64} & \textbf{0.992 $\pm$ 0.216}
& \underline{10.55 $\pm$ 2.20} & \underline{1.329 $\pm$ 0.607}
& 43.35 $\pm$ 5.82 & 1.591 $\pm$ 0.340 \\

\bottomrule
\end{tabular}}
\end{table}

\subsection{Partial Observation}

To simulate real-world scenarios where full state observation is often inaccessible, we first conduct controlled experiments on the Lorenz63 system as a benchmark for reconstructing dynamics from incomplete data. In this setting, the model is trained and evaluated using only restricted subsets of the system variables (e.g., single or dual dimensions) while the remaining dimensions act as unobserved latent states. As evidenced in Table~\ref{tab:partial_obs_smape_kld}, PhyxMamba establishes a new state-of-the-art across all partial observation configurations. Most notably, in the challenging single-variable setting (observing only $x$), our method achieves an sMAPE of \textbf{10.16} and a $D_{\text{stsp}}$ of \textbf{0.093}, substantially outperforming the strongest baseline, NHiTS, which yields an sMAPE of 31.78 and a higher $D_{\text{stsp}}$ of 0.133. Furthermore, while baselines like NBEATS suffer from severe topological collapse ($D_{\text{stsp}} > 10$), our method maintains high geometric fidelity. This corroborates that our physics-informed time-delay embedding reconstructs the global phase space from local observables. By engaging the Mamba backbone to unfold latent temporal dependencies and applying geometric regularization to preclude topological collapse, our method ensures robust attractor fidelity even under restricted single-view observations.

We further extend the partial observation analysis to ENSO, since multivariate climate dynamics are particularly susceptible to missing state variables. Table~\ref{tab:partial_obs_enso} reports two physically meaningful settings. The Niño 3.4 setting observes only SST and SSTA in the Niño 3.4 region, which tests whether a model can reconstruct the broader climate state from the most representative local monitoring region. The SSTA-only setting observes only SSTA across all Niño regions, which retains the anomaly signals most directly associated with ENSO phase while hiding the corresponding absolute-temperature channels. These two settings therefore evaluate complementary practical cases: localized but information-rich monitoring, and spatially broader but variable-restricted monitoring. Under the Niño 3.4 setting, PhyxMamba reduces sMAPE from 76.37 for the strongest baseline, NBEATS, to \textbf{44.49} and reduces $D_{\text{stsp}}$ from 3.498 to \textbf{0.315}. Under the SSTA-only setting, it also lowers sMAPE from 140.65 to \textbf{84.52} and $D_{\text{stsp}}$ from 0.721 to \textbf{0.243}. These results indicate that, even when only physically restricted observables are available, PhyxMamba can infer a faithful dynamical surrogate that preserves the attractor-level structure of the underlying chaotic climate system. This supports the central goal of chaotic system reconstruction: recovering long-term dynamical behavior from incomplete measurements, rather than merely improving point-wise forecasting accuracy.

\begin{table}[!t]
\centering
\caption{Performance comparison under two physically meaningful partial observation settings for ENSO.}
\label{tab:partial_obs_enso}
\setlength{\tabcolsep}{3pt}
\resizebox{\columnwidth}{!}{%
\begin{tabular}{c|>{\columncolor{gray!15}}c>{\columncolor{gray!15}}c|cc|cc}
\toprule
\multirow{2}{*}{Setting}
& \multicolumn{2}{>{\columncolor{gray!15}}c|}{\textbf{PhyxMamba}}
& \multicolumn{2}{c|}{NBEATS}
& \multicolumn{2}{c}{NHiTS} \\
\cmidrule(lr){2-3} \cmidrule(lr){4-5} \cmidrule(lr){6-7}
& sMAPE ($\downarrow$) & $D_{\text{stsp}}$ ($\downarrow$)
& sMAPE ($\downarrow$) & $D_{\text{stsp}}$ ($\downarrow$)
& sMAPE ($\downarrow$) & $D_{\text{stsp}}$ ($\downarrow$) \\
\midrule

Niño 3.4
& \textbf{44.49 $\pm$ 6.42} & \textbf{0.315 $\pm$ 0.138}
& \underline{76.37 $\pm$ 3.52} & \underline{3.498 $\pm$ 0.062}
& 115.46 $\pm$ 5.74 & 9.899 $\pm$ 0.728 \\

SSTA-only
& \textbf{84.52 $\pm$ 7.97} & \textbf{0.243 $\pm$ 0.028}
& \underline{140.65 $\pm$ 4.01} & \underline{0.721 $\pm$ 0.216}
& 156.13 $\pm$ 5.00 & 1.095 $\pm$ 0.120 \\

\bottomrule
\end{tabular}}
\end{table}

\subsection{Sampling Granularity and Context Length Analysis}
\begin{figure}
    \centering
    \includegraphics[width=\linewidth]{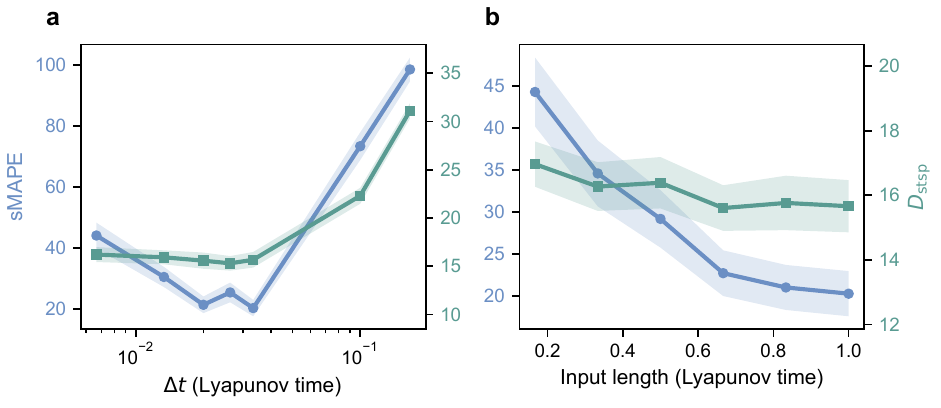}
    \caption{Chaotic system reconstruction quality on the Lorenz96 system as a function of (a) time resolution and (b) input sequence length.}
    \label{fig:lorenz96_resolution}
\end{figure}
We investigate the impact of temporal sampling granularity and historical context length on reconstruction quality using the Lorenz96 system, as illustrated in Figure~\ref{fig:lorenz96_resolution}.

\subsubsection{\textbf{Sampling Granularity}}
Figure~\ref{fig:lorenz96_resolution}(a) studies the effect of temporal resolution. We fix the input sequence length at 30 steps while varying the sampling step size from approximately $1/150$ to $1/6$ Lyapunov times. The results demonstrate that PhyxMamba maintains robust performance across a wide range of temporal resolutions. However, performance deteriorates sharply when the step size becomes excessively large, since the trajectory can undergo complex evolution within a single time step, making the deterministic dynamics indistinguishable from stochastic noise. Conversely, at extremely small step sizes, the span of the input window becomes insufficient to uniquely localize the current state within the attractor, leading to a decline in sMAPE. Notably, the long-term geometric fidelity ($D_{\text{stsp}}$) remains stable, indicating that the model successfully captures the invariant physical laws regardless of initial conditions.

\subsubsection{\textbf{Observation Context Length}}
Figure~\ref{fig:lorenz96_resolution}(b) examines the effect of available historical context. We fix the resolution at $1/30$ Lyapunov time while varying the input length. While sMAPE degrades with shorter observation windows, $D_{\text{stsp}}$ remains remarkably resilient. This confirms that PhyxMamba serves as a valid dynamical surrogate to capture the global geometry of the system attractor.

\subsection{Robustness Analysis}
\begin{figure}
    \centering
    \includegraphics[width=\linewidth]{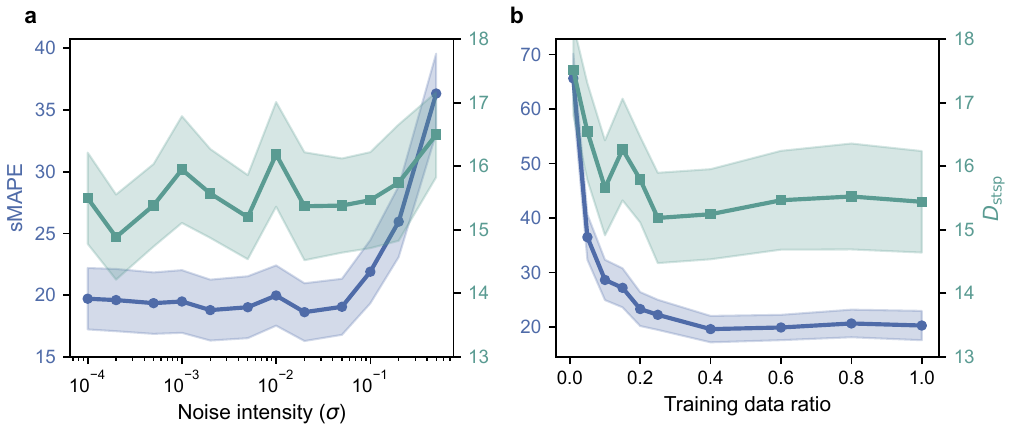}
    \caption{Chaotic system reconstruction quality on the Lorenz96 system as a function of (a) noise intensity and (b) training data ratio.}
    \label{fig:noise_ratio}
 
\end{figure}
\subsubsection{\textbf{Robustness against Noise}}
To evaluate the resilience of PhyxMamba against observational uncertainty, we introduce zero-mean Gaussian noise $\mathcal{N}(0, \sigma^2)$ into the trajectories of the Lorenz96 system during both training and inference, varying the intensity $\sigma$ from $10^{-4}$ to $5\times 10^{-1}$. As illustrated in Figure~\ref{fig:noise_ratio}a, the short-term point-wise accuracy (sMAPE) deteriorates as noise intensity increases, particularly when $\sigma > 10^{-1}$. It is theoretically expected, as the superposition of stochastic noise onto chaotic signals inherently obscures the precise trajectory coordinates, making exact matching intractable. However, $D_{\text{stsp}}$ remains remarkably stable even under significant noise contamination, indicating that the predicted trajectories continue to reside on the correct manifold. It suggests that PhyxMamba functions as an effective surrogate, successfully internalizing the invariant statistical structures of the chaotic system rather than overfitting to noise-corrupted variations.

\subsubsection{\textbf{Robustness against Training Data Ratio}}
We further investigate the data efficiency of our framework by training the model on varying subsets of the historical data from the Lorenz96 system, ranging from $1\%$ to $100\%$. The results summarized in Figure~\ref{fig:noise_ratio}b indicate a rapid convergence in model performance. Both short-term prediction accuracy and long-term geometric fidelity stabilize once the training data ratio reaches approximately $20\%$. Complete numerical results with confidence intervals for the robustness analyses are provided in Appendix~\ref{app:app_robustness}. This high data efficiency underscores the strength of our designs, which enables the extraction of underlying dynamics from limited trajectory samples.

\subsection{Complexity Analysis} \label{sec:complexity}
We report trainable parameter count and inference time for an autonomous rollout of length $10T_L$, using the same trainable baseline groups as Table~\ref{tab:simulation}.
\begin{table*}[t]
\centering
\caption{Model complexity analysis on representative simulated and real-world datasets. Inference time is reported as mean $\pm$ standard deviation for an autonomous rollout of length $10T_L$.}
\label{tab:complexity}
\setlength{\tabcolsep}{3pt}
\begin{adjustbox}{width=\textwidth,center}
\begin{tabular}{ccccccccc}
\toprule
\multirow{2}{*}{Method}
& \multicolumn{2}{c}{Lorenz63}
& \multicolumn{2}{c}{Rössler}
& \multicolumn{2}{c}{Lorenz96}
& \multicolumn{2}{c}{ECG} \\
\cmidrule(lr){2-3}\cmidrule(lr){4-5}\cmidrule(lr){6-7}\cmidrule(lr){8-9}
& Param (M) & Time (ms)
& Param (M) & Time (ms)
& Param (M) & Time (ms)
& Param (M) & Time (ms) \\
\midrule
\rowcolor{gray!15}
\textbf{PhyxMamba}
& \textbf{3.65} & \textbf{183.76 $\pm$ 2.38}
& \textbf{0.97} & \textbf{106.12 $\pm$ 3.42}
& \textbf{2.76} & \textbf{285.19 $\pm$ 2.67}
& \textbf{3.63} & \textbf{935.39 $\pm$ 28.51} \\
\midrule
NBEATS
& 6.24 & 2052.07 $\pm$ 285.74
& 6.24 & 2418.02 $\pm$ 830.10
& 6.24 & 2051.68 $\pm$ 118.35
& 6.24 & 2183.45 $\pm$ 130.93 \\
NHiTS
& 7.17 & 1599.60 $\pm$ 263.96
& 7.17 & 1637.66 $\pm$ 326.75
& 7.17 & 1625.34 $\pm$ 303.85
& 7.17 & 1694.81 $\pm$ 39.67 \\
CrossFormer
& 7.56 & 176.62 $\pm$ 13.28
& 7.56 & 174.77 $\pm$ 12.27
& 7.57 & 173.80 $\pm$ 12.16
& 7.57 & 180.66 $\pm$ 13.18 \\
PatchTST
& 6.04 & 33.13 $\pm$ 1.37
& 6.04 & 32.98 $\pm$ 1.15
& 6.04 & 33.25 $\pm$ 1.16
& 6.07 & 35.74 $\pm$ 1.33 \\
TimesNet
& 143.01 & 1105.63 $\pm$ 6.52
& 143.01 & 1118.16 $\pm$ 4.80
& 143.02 & 1118.32 $\pm$ 3.22
& 143.02 & 1180.06 $\pm$ 7.66 \\
TiDE
& 0.60 & 57.12 $\pm$ 0.97
& 0.60 & 57.50 $\pm$ 0.77
& 0.63 & 86.80 $\pm$ 1.12
& 0.66 & 86.29 $\pm$ 0.95 \\
iTransformer
& 4.54 & 23.93 $\pm$ 0.38
& 4.54 & 24.17 $\pm$ 0.87
& 4.54 & 39.64 $\pm$ 0.90
& 4.55 & 23.98 $\pm$ 0.68 \\
\midrule
Koopa
& 0.09 & 51.21 $\pm$ 1.22
& 0.09 & 51.31 $\pm$ 1.06
& 0.10 & 51.07 $\pm$ 0.93
& 0.10 & 51.24 $\pm$ 1.76 \\
PLRNN
& 0.37 & 855.34 $\pm$ 35.56
& 0.37 & 845.33 $\pm$ 20.60
& 0.43 & 929.98 $\pm$ 32.50
& 0.12 & 2283.17 $\pm$ 88.50 \\
PRC
& 0.003 & 2160.28 $\pm$ 57.71
& 0.003 & 2081.44 $\pm$ 40.71
& 0.003 & 3596.25 $\pm$ 55.42
& 0.003 & 793.72 $\pm$ 5.43 \\
HoGRC
& 0.003 & 1775.32 $\pm$ 10.23
& 0.003 & 1911.88 $\pm$ 5.86
& 0.003 & 2772.48 $\pm$ 40.89
& 0.003 & 594.21 $\pm$ 6.92 \\
\bottomrule
\end{tabular}
\end{adjustbox}
\end{table*}

As shown in Table~\ref{tab:complexity}, PhyxMamba maintains a practical balance between reconstruction quality and computational cost. On Lorenz96, it uses 2.76M parameters, far fewer than TimesNet (143.02M), and completes a $10T_L$ rollout in 285.19 ms, much faster than PRC (3596.25 ms) and HoGRC (2772.48 ms). The ECG results provide corresponding real-world evidence: the smaller patch size increases runtime to 935.39 ms, while PhyxMamba remains faster than NBEATS, NHiTS, and PLRNN and retains the fidelity gains reported in Table~\ref{tab:simulation}. Although several lightweight predictors have lower latency, their dynamical fidelity is substantially weaker in Table~\ref{tab:simulation}. These results indicate that PhyxMamba keeps computational cost moderate while preserving strong chaotic system reconstruction quality.

\subsection{Hyperparameter Study} \label{sec:hyperparameter-study}
\begin{figure}[t]
    \centering
    \includegraphics[width=\linewidth]{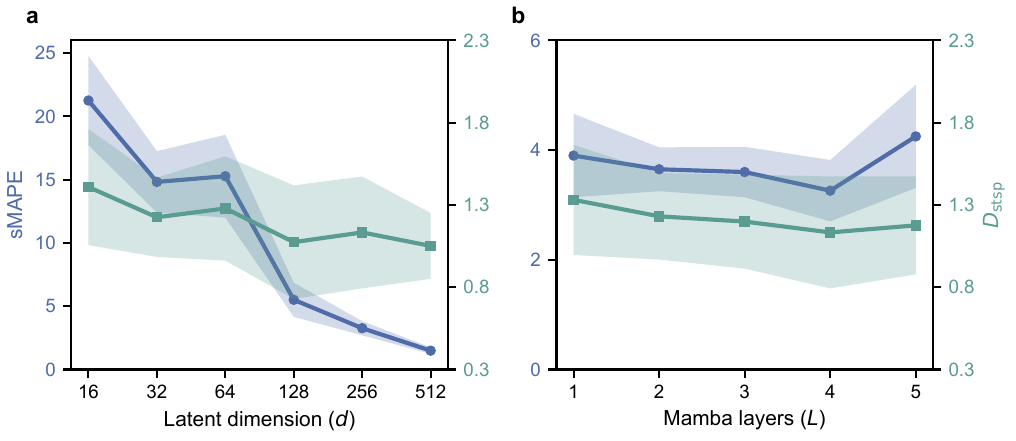}
    \caption{Lorenz63 reconstruction quality under capacity-related hyperparameters: (a) latent patch dimension $d$ and (b) number of Mamba layers $L$. Shaded regions denote 95\% confidence intervals over multiple runs from randomly sampled initial conditions.}
    \label{fig:capacity_hyperparameters}
\end{figure}

Complete hyperparameter settings are provided in Appendix~\ref{app:hyperparameter} and Table~\ref{tab:hyper_setting}. We examine capacity-related hyperparameters that directly affect chaotic system reconstruction, including the latent patch dimension $d$ and the number of Mamba layers $L$. All hyperparameter analyses are conducted on a single NVIDIA RTX 4090 GPU. Figure~\ref{fig:capacity_hyperparameters} studies two hyperparameters that directly change model capacity and parameter count, with complete numerical values and confidence intervals provided in Tables~\ref{tab:impact_d_lorenz63} and~\ref{tab:hier_with_ci_l63} of Appendix~\ref{app:hyperparameter-study}. The latent patch dimension $d$ specifies the dimensionality of the compressed patch token used for latent dynamics evolution. Increasing $d$ substantially improves short-term accuracy, reducing sMAPE from 21.25 at $d=16$ to 3.26 at $d=256$ and 1.49 at $d=512$. The attractor-level error also shows a slower overall improvement, with $D_{\text{stsp}}$ decreasing from 1.409 at $d=16$ to 1.051 at $d=512$. Since larger latent dimensions also increase memory and computation, we use $d=256$ in the main experiments as a practical capacity-efficiency trade-off. The number of Mamba layers $L$ controls the depth of residual-stacked state-space evolution. The Lorenz63 results show that increasing depth improves reconstruction up to $L=4$, which we select in the main experiments, whereas the deeper setting $L=5$ deteriorates performance, with sMAPE reaching 4.25. These results indicate that capacity should be increased only to the point where it improves system reconstruction, since excessive latent dimensionality or depth can add parameters without clear fidelity gains. Additional analyses of the time-delay embedding parameters $(m,\tau)$, patch size $D$, multi-patch evolutionary guidance steps $M$, and guidance weight $\lambda_p$ are reported in Appendix~\ref{app:hyperparameter-study}.

% \textbf{Robustness against Input Observation Lengths.}
% We explore the effect of input observation lengths to our method.

% \subsection{test}
% \input{Tables/Chaotic_compact}
% \input{Tables/Real_full}
% \input{Tables/dummy}

% \begin{figure*}[htbp]
%     \centering
%     \includegraphics[width=\textwidth]{Figures/robustness.pdf}
%     \caption{Model robustness against noise and training data on Rössler system and EEG dataset.}
%     \label{fig:noise_ratio}
% \end{figure*}

% \begin{figure}
%     \centering
%     \includegraphics[width=\textwidth]{Figures/noise.pdf}
%     \caption{Caption}
%     \label{fig:noise}
% \end{figure}

% \begin{figure}
%     \centering
%     \includegraphics[width=\textwidth]{Figures/data_ratio.pdf}
%     \caption{Caption}
%     \label{fig:enter-label}
% \end{figure}

% \centering
% \includegraphics[width=0.8\linewidth]{Figures/noise_analysis.pdf}

\section{Related Works} \label{sec:related}
% turbulence -> RC -> DL (RNN->专有模型）
\subsubsection{\textbf{Dynamical System Reconstruction}}
Dynamical system reconstruction has gained significant attention due to its importance across diverse scientific and engineering fields. A key breakthrough is reservoir computing (RC)~\cite{gauthier2021next,srinivasan2022parallel,li2024higher}, which uses a fixed, randomly initialized recurrent neural network (RNN) to map inputs into a high-dimensional dynamical space, training only the readout layer for efficient learning. More recently, deep recurrent models have been used to capture temporal dependencies in nonlinear systems. To mitigate gradient explosion during chaotic trajectory modeling, recent works~\cite{hess2023generalized,mikhaeil2022difficulty} employ teacher forcing, substituting internal states with ground truth during training. Long-horizon predictive learners further use recurrent curricula to extract dynamics from context frames~\cite{wang2023predrnn}, while observer-theoretic architectures introduce dynamical regularization for one- and multi-step forecasts~\cite{liang2025observer}. Additionally, physics-informed regularizations based on optimal transport and Maximum Mean Discrepancy (MMD) with neural operators~\cite{jiang2023training,schiff2024dyslim} help preserve chaotic attractors' invariant measures. Generic forecasting architectures~\cite{nie2022time,liu2023itransformer,zhang2023crossformer,ekambaram2023tsmixer,liu2024gph} are not designed to jointly ensure that a short context localizes the current state on the attractor and that autonomous rollouts preserve its long-run invariant measure. PhyxMamba addresses these coupled requirements by integrating attractor manifold representation, multi-step generative learning, and invariant-measure alignment within a unified state-space framework.

\subsubsection{\textbf{Mamba in Time Series Modeling}}
Mamba~\cite{gu2023mamba,DBLP:conf/icml/DaoG24}, rooted in state space models, has recently gained prominence in sequence modeling across fields like natural language processing and computer vision. In time series analysis, representative variants such as S-Mamba~\cite{wang2025mamba} adapt Mamba blocks to temporal and channel-wise dependency modeling. These methods primarily use Mamba as an encoder for correlation modeling, without explicitly learning intrinsic dynamical processes. Our approach differs by embedding Mamba blocks in a generative framework, achieving effective reconstruction of the system’s intrinsic dynamics.
% Mamba~\cite{gu2023mamba,DBLP:conf/icml/DaoG24}, originating from state space models, has emerged as a novel architecture for sequence modeling, garnering significant interest across domains such as natural language processing and computer vision. Within the realm of time series modeling, there is an increasing trend to adopt Mamba architectures to effectively capture temporal and channel-wise correlations. For instance, TimeMachine~\cite{ahamed2024timemachine} and DTMamba~\cite{wu2024dtmamba} utilize parallel Mamba modules operating at low and high resolutions, respectively, to capture both local and global contextual information. S-Mamba~\cite{wang2025mamba} further extends this by incorporating bi-directional Mamba layers for modeling inter-channel correlations. Other works seek to synergize the strengths of Mamba with complementary modules; Mambaformer~\cite{xu2024integrating} leverages Mamba to learn long-term dependencies while employing self-attention layers to model short-term interactions, and Bi-Mamba+~\cite{liang2024bi} enhances the Mamba block with an additional forget gate, proposing Mamba+ for improved preservation of long-term information. However, these existing approaches primarily utilize Mamba as an encoder focused on correlation modeling, without explicitly learning the underlying intrinsic dynamical processes. In contrast, our approach integrates Mamba blocks within a generative framework, enabling not only competitive forecasting performance but also effective capture of the system’s intrinsic dynamical behavior.

\section{Conclusions} \label{sec:conclusion}
This paper studies chaotic system reconstruction from short context observations. We propose PhyxMamba, which combines time-delay phase-space representation, Mamba-based autonomous rollout, multi-step evolutionary guidance, and invariant-measure alignment to preserve both local evolution and long-term attractor structure. Experiments on simulated and real-world chaotic systems show that PhyxMamba achieves strong reconstruction accuracy and attractor fidelity, while remaining robust under partial observations, varying sampling granularity, limited training data, and observational noise. These results indicate that reconstruction-oriented state-space models can serve as faithful dynamical surrogates under limited short-context observations.
% The limitation lies in the fact that forecasting performance is inherently restricted by the representational capacity of the backbone. 

\bibliographystyle{IEEEtran}
\bibliography{main}

\vspace{-0.7em}
\begin{IEEEbiography}[{\includegraphics[width=1.0in,height=1.25in,clip,keepaspectratio]{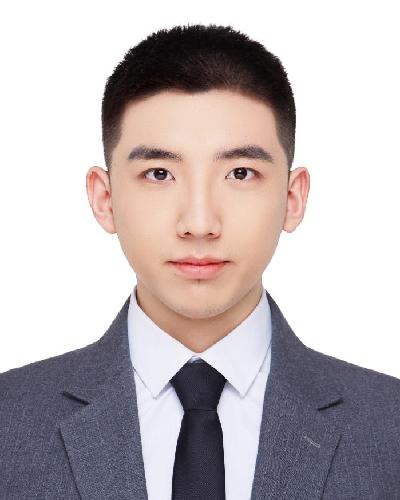}}]{Chang Liu}
received the B.S. degree in electronic engineering from Tsinghua University, Beijing, China, in 2023. He is currently working towards the Ph.D. degree with the Department of Electronic Engineering, Tsinghua University. His current research interests include dynamical systems, complex networks, and AI for science. He has publications in journals and conferences such as Nature Communications, ICML, KDD, and WWW.
\end{IEEEbiography}
\vspace{-1.2em}

\begin{IEEEbiography}[{\includegraphics[width=1.0in,height=1.25in,clip,keepaspectratio]{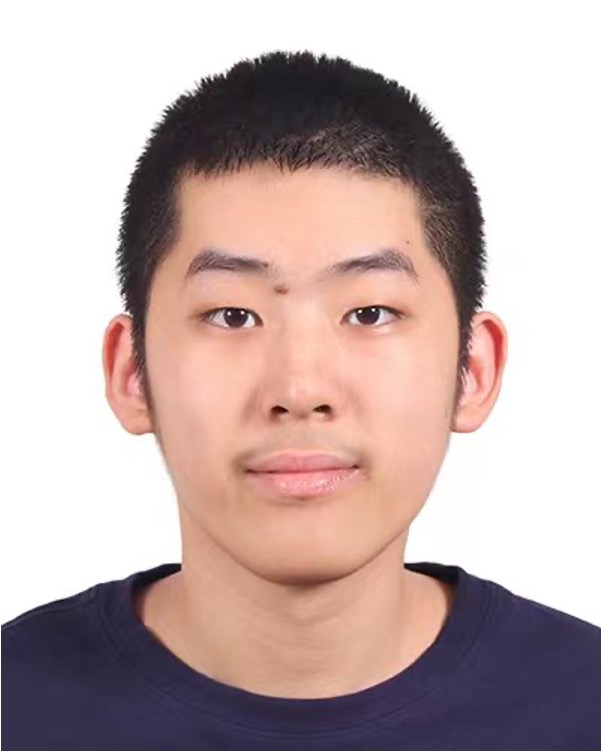}}]{Bohao Zhao}
is currently pursuing the B.S. degree in the Department of Electronic Engineering at Tsinghua University, Beijing, China. His primary research interests lie in AI for science, particularly in the modeling of nonlinear dynamical systems.
\end{IEEEbiography}
\vspace{-1.2em}

\begin{IEEEbiography}[{\includegraphics[width=1.0in,height=1.25in,clip,keepaspectratio]{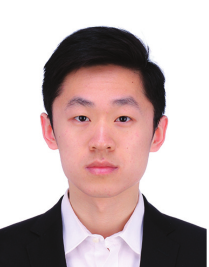}}]{Jingtao Ding}
received the B.S. degree in electronic engineering and the Ph.D. degree in electronic engineering from Tsinghua University, Beijing, China, in 2015 and 2020, respectively. He is currently a Post-Doctoral Research Fellow with the Department of Electronic Engineering, Tsinghua University. His research interests include mobile computing, spatiotemporal data mining, and user behavior modeling. He has more than 30 publications in journals and conferences such as TKDE, TOIS, KDD, NeurIPS, WWW, ICLR, SIGIR, and IJCAI.
\end{IEEEbiography}
\vspace{-1.2em}

\begin{IEEEbiography}[{\includegraphics[width=1.0in,height=1.25in,clip,keepaspectratio]{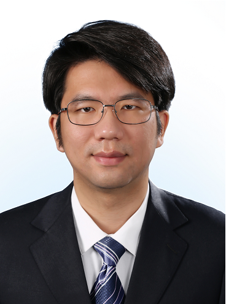}}]{Huandong Wang}
(Member, IEEE) received the B.S. degree in electronic engineering and the second B.S. degree in mathematical sciences from Tsinghua University, Beijing, China, in 2014 and 2015, respectively, and the Ph.D. degree in electronic engineering from Tsinghua University, Beijing, China, in 2019. He is currently a research associate with the Department of Electronic Engineering, Tsinghua University. His research interests include urban computing, mobile Big Data mining, and machine learning.
\end{IEEEbiography}
\vspace{-1.2em}

\begin{IEEEbiography}[{\includegraphics[width=1.0in,height=1.25in,clip,keepaspectratio]{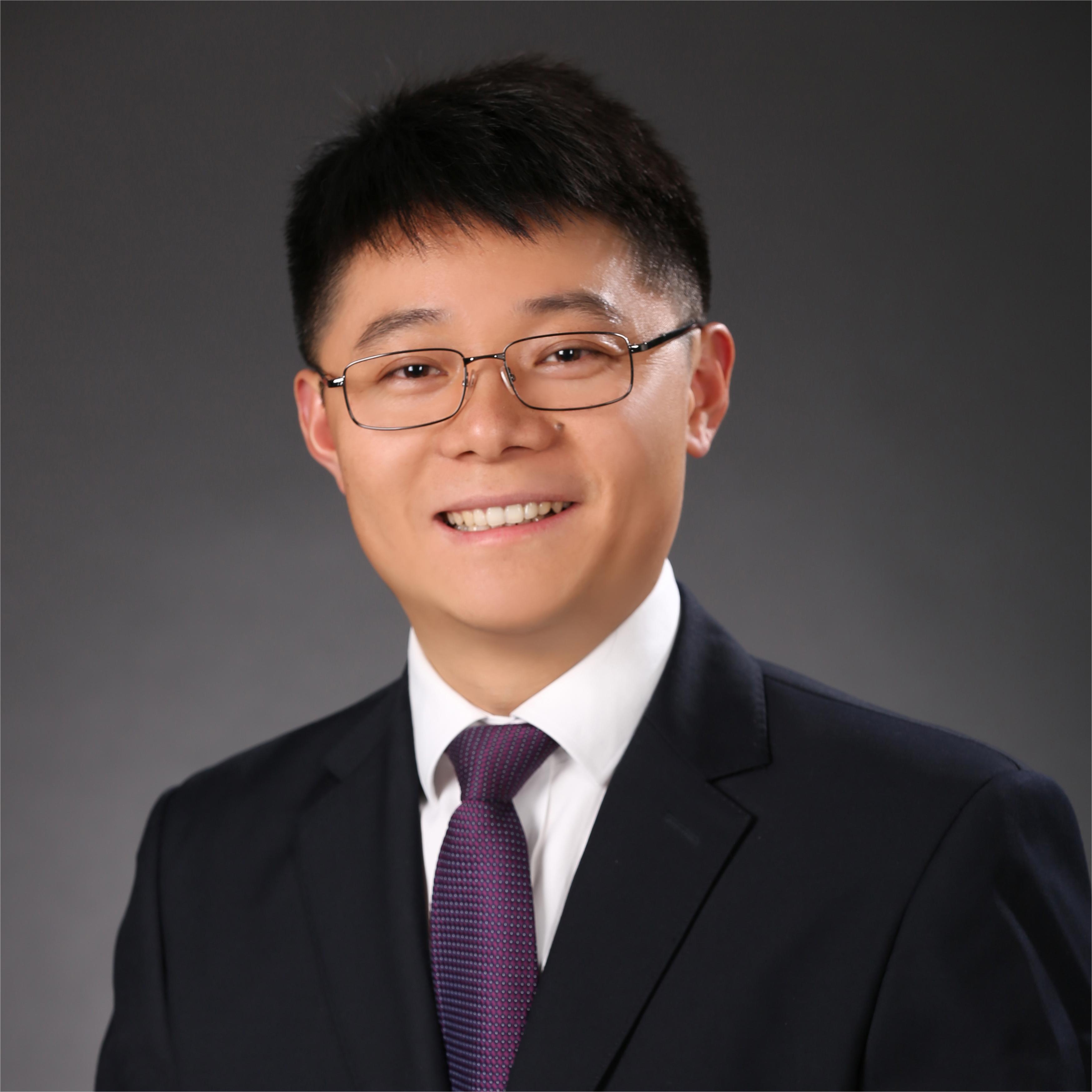}}]{Yong Li}
(M'09-SM'16) received the B.S. degree in electronics and information engineering from Huazhong University of Science and Technology, Wuhan, China, in 2007, and the Ph.D. degree in electronic engineering from Tsinghua University, Beijing, China, in 2012. He is currently a professor with the Department of Electronic Engineering, Tsinghua University.
Prof. Li has served as General Chair, TPC Chair, and SPC/TPC Member for several international workshops and conferences, and he is on the editorial board of two IEEE journals. His papers have received more than 27000 citations. Among them, ten are ESI Highly Cited Papers in Computer Science, and four received conference Best Paper or Best Paper Runner-Up Awards. He received the IEEE 2016 ComSoc Asia-Pacific Outstanding Young Researchers, the Young Talent Program of China Association for Science and Technology, and the National Youth Talent Support Program.
\end{IEEEbiography}

\let\appendix\appendices
\newpage
\appendix
% \section{Appendix}
\section{Implementation Details}
In this section, we demonstrate implementation details of our methods.
\subsection{Representation}\label{app:represent}
\textbf{Takens' Embedding Theorem.} It is a key result in nonlinear dynamics, which states that the state space of a smooth dynamical system can be reconstructed from time-delay embeddings of a single observed scalar time series. Given a scalar measurement function $h: \mathcal{M}\rightarrow \mathbb{R}$ on a compact manifold $\mathcal{M}$ representing the system’s state space, the delay embedding map is defined as
\begin{equation}
    \Phi(\bm{x})=(h(\bm{x}), h(f(\bm{x})), h(f^2(\bm{x})), \dots, h(f^{m-1}(\bm{x}))),
\end{equation}
where $f:\mathcal{M}\rightarrow \mathcal{M}$ is the system’s evolution map and $m$ is the embedding dimension. Takens showed that for a generic choice of $h$, if $m>2d$ (with $d$ being the dimension of $\mathcal{M}$), the map $\Phi$ is an embedding, meaning it is a diffeomorphism onto its image, thus preserving the topology of the original system.

\textbf{Embedding-parameter selection.}
Section~\ref{sec:method-rep} gives the time-delay construction and the criteria used to select the embedding dimension $m$ and delay $\tau$. In implementation, we estimate the average mutual information over candidate delays and choose the first local minimum as $\tau$. We then increase $m$ until the correlation-integral statistic stabilizes, indicating that the reconstructed neighborhood structure has been sufficiently unfolded. A single pair $(m,\tau)$ is used for all variables within the same dataset so that the resulting multivariate delay representation has a consistent tensor shape. The selected values for all datasets are reported in Table~\ref{tab:hyper_setting}.

\subsection{State-space Models and Mamba}\label{app:ssm_mamba}
% 介绍Mamba，以及它为什么适合这个任务
% 精细建模吸引子结构（借鉴Attraos的讲法），因此采用hierarchical mamba layers
%% Introduction of Mamba (TODO)
State-space model (SSM) is a mathematical framework derived from dynamical systems. It describes the relationships between hidden states and observation variables of the system and models their temporal behavior. SSM usually maps continuous inputs $x(t)$ to corresponding outputs $y(t)$ via the state hidden representation $h(t)$:
\begin{equation}
\begin{aligned}
\frac{dh(t)}{dt} &= \bm{A}h(t) + \bm{B}x(t), \\
y(t) &= \bm{C}h(t),
\end{aligned}
\end{equation}
where $\bm{A},\bm{B},\bm{C}$ represent the parameters of SSMs. Because of the difficulty in acquiring analytical solutions for SSMs, discretization plays an important role in facilitating analysis and solution in the discrete domain. In such a process, the input signal is first sampled at fixed time intervals, and the resulting discrete state-space model can be written as follows:
\begin{equation}
\begin{aligned}
    h_k &= \overline{\bm{A}}h_{k-1} + \overline{\bm{B}}x_k,\\
    y_k &= \bm{C}h_k,
\end{aligned}
\end{equation}
where $\overline{\bm{A}}=f_A(\Delta, \bm{A})$ and $\overline{\bm{B}}=f_B(\Delta, \bm{B})$.
Discrete rules $(f_A, f_B)$ are usually chosen as the zero-order hold (ZOH) defined as:
\begin{equation}
    \overline{\bm{A}} = \exp(\Delta\bm{A}), \qquad \overline{\bm{B}} = (\Delta\bm{A})^{-1}(\exp(\Delta\bm{A}) - \bm{I})\cdot\Delta\bm{B},
\end{equation}
where $\bm{\Delta}$ denotes the step size, and $\bm{I}$ is the identity matrix.

Traditional SSMs are implemented as time-invariant systems, where transfer parameters (e.g., $\bm{B}$ and $\bm{C}$) are identical at all time steps. This linear time-invariant (LTI) characteristic, while enabling efficient computation through methods like convolutions, restricts the model's ability to adapt its behavior based on the specific content of the input sequence.
To address this limitation, Mamba~\cite{gu2023mamba} introduces a selective scan mechanism. While the continuous state evolution matrix $\bm{A}$ is typically kept fixed and often structured (e.g., diagonal) for stability and computational reasons, other key parameters that govern the discretized SSM dynamics---$\bm{B}$, $\bm{C}$, and the discretization step size $\Delta$---are rendered input-dependent.
Specifically, these parameters are dynamically computed from the current input $x_k$, typically via lightweight linear projections or small neural networks.
Such an input-dependent formulation, which forms the cornerstone of the full Mamba architecture~\cite{gu2023mamba}, empowers the model to selectively emphasize or ignore parts of the input history by dynamically modulating its effective discrete parameters.
This allows for the robust modeling of temporal dependencies by modulating state transitions, effectively filtering irrelevant information, and adapting state transitions based on contextual cues within the input.

Building upon these principles, Mamba2~\cite{DBLP:conf/icml/DaoG24} represents a subsequent refinement, developed under the Structured State Space Duality (SSD) framework, which elucidates deeper connections between SSMs and attention mechanisms. The core of Mamba2 is its SSD layer, where the state transition dynamics $\bm{A}_k$ are simplified (often to a scalar representation per step), and the layer typically utilizes larger head dimensions (e.g., 64 or 128), aligning more closely with Transformer conventions. These design choices enable the SSD layer to achieve significant speedups, reportedly 2-8x faster than the original Mamba's selective scan, by leveraging hardware matrix multiplication units more effectively. Furthermore, the Mamba-block architecture incorporates changes aimed at improved scalability and system efficiency, such as parallelizing the initial projections that generate the SSM parameters ($\bm{A}_k$, $\bm{B}_k$, $\bm{C}_k$) and the primary input $x_k$. While preserving Mamba's foundational principles of input-dependent selectivity and linear-time processing characteristic, Mamba2 aims to deliver these capabilities with superior computational performance and enhanced amenability to large-scale model training paradigms.
We adopt Mamba2 as the backbone unless otherwise specified, referring to it as Mamba for brevity.
% Given these advancements, particularly in efficiency and scalability, Mamba2 is the variant employed in this work.

\subsection{Maximum Mean Discrepancy Regularization} \label{app:mmd}
To faithfully reproduce key statistical invariants and attractor geometry, we incorporate regularization terms designed to preserve the distribution of long-term system statistics. According to existing works~\cite{schiff2024dyslim}, such a regularization term should satisfy the following properties: (i) it should respect the underlying geometry of the space $\mathcal{A}$ and allow for comparison between measures even when their supports do not overlap; (ii) it must admit an unbiased estimator that can be computed from samples; (iii) it should have low computational complexity relative to both the system’s dimensionality and the number of samples; (iv) it needs to guarantee convergence properties on the space of measures defined over $\mathcal{A}$, meaning that if $D(p_1, p) \rightarrow 0$, then $p_1$ converges to $p$; and (v) it should achieve parametric rates of estimation, so that the sampling error $|D-\hat{D}|$ does not depend on the dimension of the system.

Integral Probability Metrics (IPMs) provides a general solution for distribution comparison. For any two distributions $p_1$ and $p_2$, IPMs are defined as follows:
\begin{equation}
    \text{IPM}(p_1, p_2) = \text{sup}_{\kappa \in \mathcal{K}}|\mathbb{E}_{\bm{u}\sim p_1}[\kappa(\bm{u})] - \mathbb{E}_{\bm{u}'\sim p_2}[\kappa(\bm{u}')]|,
\end{equation}
where $\mathcal{K}$ denotes the specific function class. If the function space $\mathcal{K}$ is rich enough, $\text{IPM}(p_1, p_2) \rightarrow 0$ indicates $p_1 \rightarrow p_2$. Maximum mean discrepancy (MMD), a member of this class, has gained significant attention due to its favorable theoretical and computational properties and ability to satisfy the requirements outlined above. MMD measures the distance between two probability distributions by comparing their embeddings. Specifically, given two distributions $p_1$ and $p_2$ in a reproducing kernel Hilbert space (RKHS). Specifically, given two distributions, the MMD is defined as:
\begin{equation}
    \mathrm{MMD}(p_1, p_2) \coloneqq \text{sup}_{||f||_{\mathcal{H} \leq 1}}|\mathbb{E}_{\bm{u} \sim p_1}[f(\bm{u})] - \mathbb{E}_{\bm{u}' \sim p_2}[f(\bm{u}')]|,
\end{equation}
where $\mathcal{H}$ is an RKHS, and $f(\cdot)$ ranges over functions in $\mathcal{H}$ with norm at most one. This formulation allows for an unbiased estimator that can be efficiently computed from samples drawn from $p_1$ and $p_2$. Moreover, the MMD respects the geometry of the underlying space and can effectively compare distributions even when their supports do not overlap.
Using the reproducing property of $\mathcal{H}$ and the Riesz representation theorem, MMD can be represented as follows:
\begin{equation}
\begin{aligned}
    \mathrm{MMD}^2(p_1, p_2)
    &= \mathbb{E}_{\bm{u}, \bm{u'} \sim p_1}[\kappa(\bm{u}, \bm{u'})]
     + \mathbb{E}_{\bm{v}, \bm{v'} \sim p_2}[\kappa(\bm{v}, \bm{v'})] \\
    &\quad - 2\mathbb{E}_{\bm{u}\sim p_1, \bm{v}\sim p_2}[\kappa(\bm{u}, \bm{v})].
\end{aligned}
\end{equation}
where $\kappa: \mathcal{A} \times \mathcal{A} \rightarrow \mathbb{R}$ is a kernel function. For the implementation of kernel function $\kappa$, we employ a mixture of rational quadratic kernels following the existing works~\cite{schiff2024dyslim}, formulated as:
\begin{equation}
    \kappa_{\bm{\sigma}}(\bm{u}, \bm{u}') = \sum_{\sigma_q \in \bm{\sigma}} \kappa_{\sigma_q}(\bm{u}, \bm{u}') = \sum_{\sigma_q \in \bm{\sigma}}\frac{\sigma_q^2}{\sigma_q^2 + ||\bm{u} - \bm{v}||_2^2},
\end{equation}
where $\bm{\sigma} = \{0.2, 0.5, 0.9, 1.3\}$, which is also consistent with existing works.

\section{Experimental Details}
\subsection{Details of Datasets} \label{app:dataset}
The canonical Lorenz63, Rössler, and Lorenz96 systems and their parameter settings are defined in Section~\ref{sec:exp-setup}. This appendix focuses on dataset construction details and additional synthetic systems used for supplementary evaluation.

\noindent \textbf{Forced FitzHugh-Nagumo.} The Forced FitzHugh-Nagumo system is formulated as:
\begin{align}
    \frac{dx}{dt} &= a + x^2y - (b+1)x + f \cos(z) \\ \frac{dy}{dt} &= bx - x^2y \\ \frac{dz}{dt} &= \omega
\end{align}
The parameter values are $a=0.7, b=0.8,f=0.40, \omega=0.044$, which can produce a strange attractor.

\noindent \textbf{Hindmarsh-Rose.} The Hindmarsh-Rose system is formulated as
\begin{align}
    \frac{dx}{dt} &= \frac{1}{\tau_x} (y - ax^3 + bx^2 + z) - x \\
\frac{dy}{dt} &= z - ax^3 - (d-b)x^2 \\
\frac{dz}{dt} &= \frac{1}{\tau_z} (c - sx - z)
\end{align}
The parameter values are $a=0.49, b=1.0, c=0.0322, d=1.0, s=1.0, \tau_x=0.03, \tau_z=0.8$, which can provide a strange attractor.
% Dataset construction process and evaluation settings are the same as other simulated systems (\textit{i.e.}, Lorenz63, Rössler, Lorenz96).

The additional synthetic systems follow the same numerical integration, Lyapunov-time resampling, training-segment construction, and testing protocol as the canonical simulated systems described in Section~\ref{sec:exp-setup}.

\noindent\textbf{Electrocardiogram (ECG) dataset.}
We adopt the Electrocardiogram (ECG) dataset from existing works~\cite{mikhaeil2022difficulty}. ECG dataset records a scalar physiological (heart muscle potential) time series, and it is embedded into a 5-dimensional space using the PECUZAL algorithm~\cite{kramer2021unified}. The data contains two trajectories, and each trajectory contains 100,000 time points ($\approx 143s$). The maximum Lyapunov exponent is estimated as $\lambda_{\max}=(2.19 \pm 0.05)\frac{1}{\text{s}}$, which is consistent with the literature~\cite{mikhaeil2022difficulty}. We resample two trajectories with an interval of 10 time steps, and the granularity of the resulting trajectory is about 30 time points per Lyapunov time. We use one of the trajectories to construct the training and validation datasets, following the same split protocol as the simulated systems.
We split another trajectory with a sliding window of 500 time steps. For each sliding window, the last 300 time steps (about 10 Lyapunov times) are treated as reconstruction objectives. The model takes the preceding 30 time steps (about 1 Lyapunov time) as the historical observation context. The remaining 170 time steps are treated as intervals to ensure the difference between test samples. After processing, we derive 20 test samples.

\noindent\textbf{Worm dataset.} It consists of motion recordings from 12 \textit{Caenorhabditis elegans} worms in a foraging context without a food source. The worms are placed onto a 9.1-cm assay plate. Their movement was confined to the plate by a copper ring with a 5-cm radius that was pressed into the agar surface. After a 5-minute acclimation period, each worm's behavior was recorded for 35 minutes. The recordings were made at an original frequency of 32 Hz and were subsequently downsampled to 16 Hz for the analysis. The shape of worm is initially described with a 100 dimension vector. It is then mathematically reduced to the five most significant components of shape variation, called "eigenworms," which capture the majority of the worm's movement. First, the dataset is preprocessed by excluding two worms whose recordings contained NaN values. From the remaining cohort, a training set is constructed by randomly selecting 8 worms. For each of these individuals, 6,250 time points were extracted from their respective movement recordings, yielding a total of 50,000 time points. Subsequently, a test set was compiled from the final two worms by extracting 20,000 time points from their collective recordings. We split the test trajectory with a sliding window of 400 time steps. For each sliding window, the last 300 time steps (about 10 Lyapunov times) are treated as reconstruction objectives, and the preceding 30 time steps (about 1 Lyapunov times) are treated as the historical observation context.

\noindent\textbf{ENSO dataset.} ENSO dataset provides a weekly time-series record of oceanic indicators crucial for monitoring the El Niño-Southern Oscillation (ENSO) from September 1981 to October 2022, comprising 2,145 data points. The data is structured with a single temporal dimension (weekly intervals) and contains eight variables corresponding to four key equatorial Pacific regions: Niño 1+2, Niño 3, Niño 3.4, and Niño 4. For each region, two specific variables are provided: 1) the absolute Sea Surface Temperature (SST) in degrees Celsius, and 2) the Sea Surface Temperature Anomaly (SSTA), which represents the deviation from the long-term climatological mean. We partition the data chronologically, utilizing the initial 80\% of the time points as the training set and reserving the subsequent 20\% as the test set. We split the test trajectory with a sliding window of 330 time steps. For each sliding window, the last 300 time steps (about 10 Lyapunov times) are treated as reconstruction objectives, and the preceding 30 time steps (about 1 Lyapunov times) are treated as the historical observation context.

\section{Extended Results}
\subsection{Detailed Results of Overall Performance} \label{app:overall}
\begin{figure*}[ht]
    \centering
    \includegraphics[width=\textwidth]{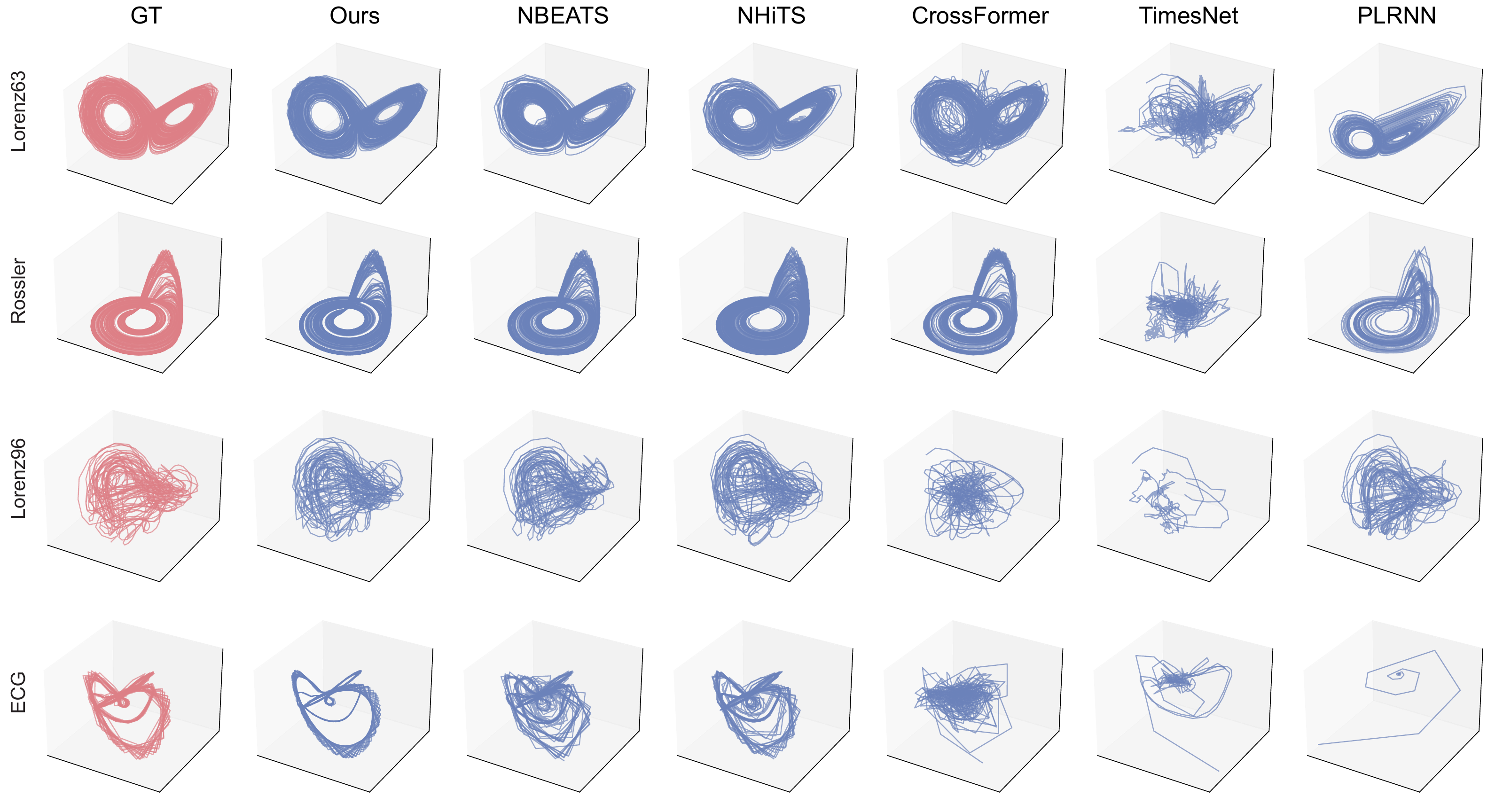}
    \caption{Visualization of reconstruction results on Lorenz63, Rössler, Lorenz96, and ECG system. We reduce the dimensionality of Lorenz96 and ECG to three dimensions using Principal Component Analysis (PCA) to facilitate this visualization.}
    \label{fig:all_reconstruction}
\end{figure*}

Here, we present a full table of model performance on simulated and real-world datasets, including results from a 95\% confidence interval, in Table~\ref{tab:full_overall}.
We also present the visualization results for the Lorenz63, Rössler, Lorenz96, and ECG datasets in Figure~\ref {fig:all_reconstruction}.

\setlength\rotFPtop{0pt plus 1fil} \begin{sidewaystable*}[p] \centering \caption{Overall performance on both simulated and real-world datasets. Reported values represent the mean ± 95\% confidence interval (CI), where CIs are computed based on multiple runs from different randomly sampled initial conditions to capture variability in the model’s predictions. The best performance of each metric is marked in \textbf{bold}, and the second-best performance is \underline{underlined}.} \label{tab:full_overall} \scriptsize \setlength{\tabcolsep}{1.2pt} \renewcommand{\arraystretch}{1.1} \resizebox{\textheight}{!}{\begin{tabular}{c|c|c|ccccccc|ccccc|ccccc} \toprule \multicolumn{2}{c|}{Category}
& \multicolumn{1}{c|}{Ours}
& \multicolumn{7}{c|}{General time series forecasting model}
& \multicolumn{5}{c|}{Physics-informed dynamical system model} & \multicolumn{5}{c}{Zero-shot foundation model}\\
\midrule
System & \diagbox{Metric}{Model} & \textbf{PhyxMamba} & NBEATS & NHiTS & CrossFormer & PatchTST & TimesNet & TiDE & iTransformer & Koopa & PLRNN & nVAR & PRC & HoGRC & Panda & Chronos2 & Chronos2-S & DynaMix & Parrot \\ \midrule \multirow{4}{*}{Lorenz63} & sMAPE ($\downarrow$) & \textbf{3.26$\pm$0.56} & \underline{5.34$\pm$0.81} & 13.93$\pm$4.21 & 13.62$\pm$1.67 & 64.39$\pm$6.40 & 35.84$\pm$4.94 & 84.88$\pm$6.99 & 67.84$\pm$6.55 & 60.68$\pm$5.75 & 59.26$\pm$6.79 & 94.32$\pm$4.32 & 18.34$\pm$3.61 & 18.82$\pm$2.05 & 105.71$\pm$6.25 & 92.18$\pm$5.39 & 93.26$\pm$5.96 & 109.60$\pm$6.31 & 112.61$\pm$5.38 \\ & $D_{\text{frac}}$ ($\downarrow$) & \textbf{0.060$\pm$0.008} & 0.075$\pm$0.008 & \underline{0.067$\pm$0.012} & 0.090$\pm$0.012 & 0.227$\pm$0.024 & 0.303$\pm$0.029 & 0.696$\pm$0.010 & 0.638$\pm$0.016 & 0.333$\pm$0.017 & 0.270$\pm$0.014 & 0.792$\pm$0.009 & 0.578$\pm$0.049 & 0.089$\pm$0.011 & 0.157$\pm$0.014 & 0.123$\pm$0.017 & 0.147$\pm$0.020 & 0.326$\pm$0.024 & 0.136$\pm$0.025 \\ & $D_{\text{stsp}}$ ($\downarrow$) & \textbf{1.133$\pm$0.340} & \underline{1.586$\pm$0.354} & 1.648$\pm$0.172 & 1.591$\pm$0.321 & 61.646$\pm$3.582 & 32.343$\pm$3.371 & 92.375$\pm$5.476 & 62.434$\pm$3.378 & 58.768$\pm$2.873 & 12.970$\pm$5.922 & 61.331$\pm$2.546 & 29.389$\pm$2.550 & 16.761$\pm$4.533 & 8.229$\pm$0.337 & 8.956$\pm$0.486 & 9.196$\pm$0.446 & 10.110$\pm$0.607 & 5.150$\pm$0.786 \\ & $D_{\text{lyap}}$ ($\downarrow$) & \textbf{0.198$\pm$0.033} & \underline{0.222$\pm$0.043} & 0.255$\pm$0.041 & 0.307$\pm$0.042 & 1.203$\pm$0.045 & 1.124$\pm$0.052 & 1.901$\pm$0.041 & 1.675$\pm$0.039 & 1.462$\pm$0.045 & 1.024$\pm$0.067 & 2.111$\pm$0.043 & 1.117$\pm$0.037 & 1.089$\pm$0.043 & 0.937$\pm$0.041 & 0.785$\pm$0.049 & 0.827$\pm$0.047 & 0.540$\pm$0.088 & 0.556 $\pm$ 0.070\\ \midrule \multirow{4}{*}{Rössler} & sMAPE ($\downarrow$) & \textbf{2.84$\pm$0.18} & \underline{14.88$\pm$0.59} & 39.90$\pm$1.56 & 27.11$\pm$1.25 & 89.16$\pm$3.45 & 42.48$\pm$2.87 & 98.48$\pm$3.42 & 85.85$\pm$5.04 & 67.78$\pm$3.65 & 51.36$\pm$4.48 & 94.95$\pm$3.20 & 28.38$\pm$2.75 & 31.03$\pm$1.88 & 154.48$\pm$2.29 & 87.56$\pm$3.51 & 85.74$\pm$3.96 & 160.56$\pm$2.86 & 94.59$\pm$3.56 \\ & $D_{\text{frac}}$ ($\downarrow$) & \textbf{0.049$\pm$0.008} & \underline{0.069$\pm$0.009} & 0.088$\pm$0.010 & 0.074$\pm$0.011 & 0.158$\pm$0.012 & 0.530$\pm$0.028 & 0.455$\pm$0.014 & 0.583$\pm$0.030 & 0.497$\pm$0.050 & 0.400$\pm$0.015 & 0.750$\pm$0.013 & 0.770$\pm$0.022 & 0.611$\pm$0.027 & 0.141$\pm$0.017 & 0.082$\pm$0.011 & 0.069$\pm$0.012 & 0.370$\pm$0.023 & 0.093$\pm$0.018 \\ & $D_{\text{stsp}}$ ($\downarrow$) & \textbf{0.034$\pm$0.009} & \underline{0.137$\pm$0.023} & 0.538$\pm$0.021 & 0.187$\pm$0.033 & 8.423$\pm$0.580 & 4.985$\pm$0.445 & 15.034$\pm$0.976 & 10.807$\pm$0.929 & 5.721$\pm$0.316 & 3.950$\pm$0.401 & 7.391$\pm$0.632 & 4.549$\pm$0.579 & 4.068$\pm$0.214 & 4.611$\pm$0.493 & 3.036$\pm$0.359 & 3.541$\pm$0.457 & 4.952$\pm$0.569 & 0.553$\pm$0.195 \\ & $D_{\text{lyap}}$ ($\downarrow$) & \textbf{0.016$\pm$0.003} & \underline{0.029$\pm$0.004} & 0.030$\pm$0.005 & 0.036$\pm$0.005 & 0.122$\pm$0.006 & 0.116$\pm$0.009 & 0.169$\pm$0.005 & 0.159$\pm$0.011 & 0.101$\pm$0.013 & 0.093$\pm$0.007 & 0.128$\pm$0.008 & 0.112$\pm$0.004 & 0.114$\pm$0.005 & 0.062$\pm$0.006 & 0.040$\pm$0.005 & 0.039$\pm$0.005 & 0.106 $\pm$ 0.011 & 0.039 $\pm$ 0.006 \\ \midrule \multirow{4}{*}{Lorenz96} & sMAPE ($\downarrow$) & \textbf{20.28$\pm$2.69} & \underline{37.51$\pm$4.32} & 51.37$\pm$3.98 & 56.64$\pm$2.55 & 116.63$\pm$2.67 & 95.93$\pm$2.52 & 123.41$\pm$2.71 & 126.08$\pm$2.80 & 114.74$\pm$2.51 & 89.90$\pm$4.18 & 86.04$\pm$3.98 & 36.42$\pm$2.86 & 47.75$\pm$2.33 & 138.20$\pm$2.22 & 117.57$\pm$2.47 & 122.05$\pm$2.11 & 139.52$\pm$2.90 & 135.25$\pm$3.31 \\ & $D_{\text{frac}}$ ($\downarrow$) & \textbf{0.164$\pm$0.020} & \underline{0.172$\pm$0.025} & 0.195$\pm$0.021 & 0.196$\pm$0.024 & 1.262$\pm$0.031 & 1.172$\pm$0.028 & 1.273$\pm$0.029 & 1.252$\pm$0.026 & 0.806$\pm$0.036 & 0.370$\pm$0.039 & 0.283$\pm$0.021 & 0.179$\pm$0.026 & 0.184$\pm$0.019 & 0.669$\pm$0.039 & 0.392$\pm$0.032 & 0.179$\pm$0.029 & 0.716$\pm$0.067 & 0.880$\pm$0.158 \\ & $D_{\text{stsp}}$ ($\downarrow$) & \textbf{15.440$\pm$0.799} & 18.043$\pm$1.000 & 18.132$\pm$0.727 & 36.191$\pm$1.353 & 141.306$\pm$3.114 & 102.414$\pm$5.324 & 174.493$\pm$4.897 & 170.792$\pm$5.210 & 128.611$\pm$4.192 & 24.020$\pm$1.655 & 23.537$\pm$0.727 & \underline{16.834$\pm$0.839} & 17.287$\pm$0.669 & 20.227$\pm$0.045 & 20.269$\pm$0.038 & 20.228$\pm$0.042 & 20.479$\pm$0.035 & 20.450$\pm$0.043 \\ & $D_{\text{lyap}}$ ($\downarrow$) & \textbf{0.204$\pm$0.028} & 0.233$\pm$0.040 & 0.222$\pm$0.039 & 0.493$\pm$0.041 & 1.691$\pm$0.041 & 1.386$\pm$0.043 & 1.820$\pm$0.039 & 1.147$\pm$0.042 & 1.587$\pm$0.044 & 0.394$\pm$0.061 & 2.225$\pm$0.046 & 0.216$\pm$0.045 & 0.219$\pm$0.044 & 0.543$\pm$0.041 & \underline{0.249$\pm$0.036} & 0.299$\pm$0.037 & 0.390 $\pm$ 0.053 & 0.404 $\pm$ 0.118 \\ \midrule \multirow{4}{*}{ECG} & sMAPE ($\downarrow$) & \textbf{37.14$\pm$15.75} & 50.77$\pm$21.89 & \underline{50.64$\pm$20.16} & 58.70$\pm$17.62 & 64.83$\pm$18.91 & 60.84$\pm$16.79 & 136.53$\pm$6.10 & 55.59$\pm$17.30 & 60.63$\pm$17.08 & 116.00$\pm$13.33 & 147.87$\pm$5.04 & 136.60$\pm$12.07 & 144.10$\pm$6.51 & 150.39$\pm$6.97 & 150.21$\pm$6.82 & 156.02$\pm$7.20 & 163.27$\pm$8.12 & 61.30$\pm$23.22 \\ & $D_{\text{frac}}$ ($\downarrow$) & \textbf{0.036$\pm$0.013} & \underline{0.038$\pm$0.020} & 0.078$\pm$0.012 & 0.443$\pm$0.050 & 0.218$\pm$0.048 & 0.143$\pm$0.039 & 0.219$\pm$0.021 & 0.242$\pm$0.056 & 0.245$\pm$0.070 & 0.529$\pm$0.102 & 0.642$\pm$0.020 & 0.147$\pm$0.020 & 0.184$\pm$0.025 & 0.328$\pm$0.049 & 0.407$\pm$0.070 & 0.642$\pm$0.112 & 1.018$\pm$0.577 & 0.099$\pm$0.087 \\ & $D_{\text{stsp}}$ ($\downarrow$) & \textbf{0.004$\pm$0.001} & \underline{0.034$\pm$0.005} & 0.035$\pm$0.005 & 0.156$\pm$0.035 & 0.440$\pm$0.079 & 0.682$\pm$0.100 & 0.834$\pm$0.076 & 0.439$\pm$0.082 & 0.463$\pm$0.068 & 1.592$\pm$0.934 & 0.638$\pm$0.053 & 3.265$\pm$1.720 & 4.015$\pm$0.822 & 0.566$\pm$0.094 & 0.775$\pm$0.196 & 0.715$\pm$0.112 & 1.853$\pm$1.263 & 0.144$\pm$0.133 \\ & $D_{\text{lyap}}$ ($\downarrow$) & \textbf{0.230$\pm$0.138} & 0.453$\pm$0.142 & 0.421$\pm$0.138 & 0.373$\pm$0.120 & 1.501$\pm$0.178 & 2.212$\pm$0.232 & 1.800$\pm$0.167 & 1.549$\pm$0.280 & 1.244$\pm$0.313 & 6.109$\pm$3.219 & 6.858$\pm$3.105 & 7.565$\pm$2.875 & 7.632$\pm$2.927 & 0.464$\pm$0.162 & 0.510$\pm$0.168 & 0.481$\pm$0.188 & 1.020$\pm$0.577 & \underline{0.367 $\pm$ 0.138} \\ \midrule \multirow{4}{*}{Worm} & sMAPE ($\downarrow$) & \textbf{89.51 $\pm$ 4.74} & 113.89$\pm$3.85 & 116.05$\pm$3.69 & \underline{90.18$\pm$3.94} & 113.52$\pm$4.07 & 100.34$\pm$4.63 & 120.14$\pm$3.92 & 134.22$\pm$3.89 & 110.70$\pm$4.24 & 122.01$\pm$4.01 & 111.84$\pm$3.19 & 102.45 $\pm$ 3.75 & 105.12 $\pm$ 3.64 & 144.16$\pm$4.30 & 116.59$\pm$4.39 & 114.07$\pm$4.06 & 144.74$\pm$3.12 & 123.11$\pm$5.14 \\ & $D_{\text{frac}}$ ($\downarrow$) & \textbf{0.159 $\pm$ 0.022} & 0.194$\pm$0.031 & \underline{0.184$\pm$0.031} & 0.294$\pm$0.035 & 0.974$\pm$0.041 & 0.815$\pm$0.043 & 0.915$\pm$0.041 & 0.806$\pm$0.043 & 0.909$\pm$0.041 & 0.638$\pm$0.067 & 1.051$\pm$0.041 & 0.855 $\pm$ 0.039 & 0.890 $\pm$ 0.042 & 0.622$\pm$0.050 & 0.443$\pm$0.049 & 0.401$\pm$0.051 & 1.069$\pm$0.116 & 1.230$\pm$0.226 \\ & $D_{\text{stsp}}$ ($\downarrow$) & \textbf{9.875 $\pm$ 1.760} & 15.495$\pm$3.299 & \underline{11.671$\pm$2.315} & 19.332$\pm$3.908 & 41.069$\pm$8.059 & 26.837$\pm$4.800 & 44.876$\pm$8.390 & 34.794 $\pm$ 7.525 & 36.829$\pm$6.191 & 25.474$\pm$6.735 & 24.410$\pm$3.876 & 18.550 $\pm$ 3.432 & 20.125 $\pm$ 3.533& 12.321$\pm$0.532 & 12.352$\pm$0.440 & 12.011$\pm$0.495 & 11.621$\pm$0.120 & 10.678$\pm$0.219 \\ & $D_{\text{lyap}}$ ($\downarrow$) & \textbf{0.079 $\pm$ 0.022} & 0.104$\pm$0.015 & 0.097$\pm$0.015 & 0.134$\pm$0.017 & 0.672$\pm$0.020 & 0.450$\pm$0.019 & 0.621$\pm$0.019 & 0.604$\pm$0.023 & 0.602$\pm$0.021 & 0.313$\pm$0.074 & 0.721$\pm$0.016 & 0.510 $\pm$ 0.018 & 0.545 $\pm$ 0.021 & 0.111$\pm$0.016 & 0.086$\pm$0.016 & \underline{0.082$\pm$0.014} & 0.174$\pm$0.031 & 0.209 $\pm$ 0.005 \\ \midrule \multirow{4}{*}{ENSO} & sMAPE ($\downarrow$) & \textbf{40.21 $\pm$ 5.58} & 47.02$\pm$6.27 & 72.18$\pm$2.05 & 64.78$\pm$4.23 & 51.22$\pm$5.94 & 48.25$\pm$6.88 & 53.36$\pm$6.01 & 49.63$\pm$6.06 & \underline{44.45$\pm$6.48} & 90.01$\pm$2.95 & 52.16$\pm$6.05 & 45.33 $\pm$ 6.03 & 48.90 $\pm$ 5.97 & 57.70$\pm$6.39 & 49.74$\pm$6.32 & 48.89$\pm$6.08 & 47.67$\pm$6.01 & 44.70$\pm$6.73 \\ & $D_{\text{frac}}$ ($\downarrow$) & \textbf{0.230 $\pm$ 0.018} & 0.717$\pm$0.020 & 0.734$\pm$0.004 & \underline{0.265$\pm$0.032} & 0.377$\pm$0.013 & 0.384$\pm$0.033 & 0.499$\pm$0.009 & 0.324$\pm$0.021 & 0.334$\pm$0.024 & 0.986$\pm$0.009 & 0.320$\pm$0.008 & 0.285$\pm$0.015 & 0.299$\pm$0.014 & 0.532$\pm$0.030 & 0.476$\pm$0.028 & 0.524$\pm$0.033 & 0.374$\pm$0.024 & 0.455$\pm$0.031 \\ & $D_{\text{stsp}}$ ($\downarrow$) &\textbf{ 0.451 $\pm$ 0.119} & 7.613$\pm$0.654 & 31.120$\pm$1.602 & 42.974$\pm$0.802 & 6.478$\pm$1.088 & 6.632$\pm$1.422 & 6.366$\pm$1.120 & 6.483$\pm$1.080 & 5.600$\pm$0.913 & 5.416$\pm$0.946 & 1.264$\pm$0.083 & \underline{1.150 $\pm$ 0.072} & 1.210 $\pm$ 0.089 & 4.891$\pm$0.577 & 4.371$\pm$0.502 & 4.583$\pm$0.527 & 6.481$\pm$0.656 & 6.001$\pm$0.509 \\ & $D_{\text{lyap}}$ ($\downarrow$) & \textbf{0.006 $\pm$ 0.001} & 0.013$\pm$0.001 & 0.008$\pm$0.001 & 0.013$\pm$0.001 & 0.013$\pm$0.001 & \underline{0.007$\pm$0.001} & 0.015$\pm$0.001 & 0.010$\pm$0.002 & 0.009$\pm$0.001 & 0.031$\pm$0.001 & 0.012$\pm$0.001 & 0.012$\pm$0.001 & 0.011$\pm$0.001 & 0.012$\pm$0.001 & 0.011$\pm$0.001 & 0.010$\pm$0.001 & 0.013$\pm$0.001 & 0.015 $\pm$ 0.001 \\ \midrule \bottomrule \end{tabular} } \end{sidewaystable*}

\subsection{Details of Hyperparameter Settings}\label{app:hyperparameter}
% \usepackage{graphicx}
% \usepackage{diagbox}
% \usepackage{booktabs}

\begin{table*}[t]
\centering
\caption{Hyperparameter settings of PhyxMamba on simulated and real-world datasets. TF and SF represent teacher forcing and student forcing, respectively. Hidden size, expand, and size per head are internal hyperparameters of Mamba blocks.}
\label{tab:hyper_setting}
\resizebox{\textwidth}{!}{%
\begin{tabular}{ccccccccccccccc}
\toprule
\midrule
 \diagbox{System}{Parameter}& $D$ & $d$ & $m$ & $\tau$ & $M$ & $L$ & $\lambda_p$ & $\lambda_c$ & $\lambda_\tau$ & TF-$lr$ & SF-$lr$ & Hidden size & Expand & Size per head \\ 
 \midrule
Lorenz63 & 10 & 256 & 3 & 7 & 3 & 4 & 0.1 & 1000 & 1 & 0.001 & 0.0001 & 1024 & 2 & 64 \\
Rössler & 5 & 256 & 3 & 7 & 3 & 1 & 0.1 & 1000 & 1 & 0.001 & 0.0001 & 1024 & 2 & 64 \\
Lorenz96 & 5 & 256 & 4 & 3 & 3 & 3 & 0.1 & 1000 & 1 & 0.001 & 0.0001 & 1024 & 2 & 64 \\
ECG & 2 & 256 & 4 & 6 & 3 & 4 & 0.1 & 1000 & 1 & 0.001 & 0.0001 & 1024 & 2 & 64 \\ 
Forced FitzHugh-Nagumo & 5 & 256 & 3 & 7 & 3 & 4 & 0.1 & 1000 & 1 & 0.001 & 0.0001 & 1024 & 2 & 64 \\
Hindmarsh-Rose & 5 & 256 & 7 & 9 & 3 & 4 & 0.1 & 1000 & 1 & 0.001 & 0.0001 & 1024 & 2 & 64 \\
Worm & 5 & 256 & 8 & 9 & 3 & 4 & 0.1 & 1000 & 1 & 0.001 & 0.0001 & 1024 & 2 & 64 \\
ENSO & 10 & 256 & 5 & 9 & 3 & 4 & 0.1 & 1000 & 1 & 0.001 & 0.0001 & 1024 & 2 & 64 \\

\midrule
\bottomrule
\end{tabular}%
}
\end{table*}
We demonstrate the hyperparameter settings of our model on simulated and real-world datasets in Table~\ref{tab:hyper_setting}. All experiments are completed on a single NVIDIA RTX
4090 GPU.

% \begin{figure*}[ht]
%     \centering
%     \includegraphics[width=\textwidth]{Figures/Lorenz96_reconstructions.pdf}
%     \caption{Visualization of reconstruction results on Lorenz96 system. We reduce the dimensionality of the high-dimensional system to three dimensions using Principal Component Analysis (PCA) to facilitate this visualization.}
%     \label{fig:l96_reconstruction}
% \end{figure*}

% \begin{figure*}[ht]
%     \centering
%     \includegraphics[width=\textwidth]{Figures/ECG_reconstructions.pdf}
%     \caption{Visualization of reconstruction results on ECG system. We reduce the dimensionality of the high-dimensional system to three dimensions using Principal Component Analysis (PCA) to facilitate this visualization.}
%     \label{fig:ecg_reconstruction}
% \end{figure*}

% \begin{figure*}
%     \centering
%     \includegraphics[width=\linewidth]{Figures/publication_trajectories_lyapunov_time_integer_ticks.pdf}
%     \caption{Dimension-wise comparison between our predictions and the ground truth for all systems.}
%     \label{fig:dimension-wise}
% \end{figure*}

\subsection{Detailed Results on Additional Synthetic Datasets} \label{sec:addition_synthetic}

\begin{table*}[t]
\centering
\caption{Overall performances on simulated FitzHugh-Nagumo and Hindmarsh-Rose datasets. Reported values represent the mean ± 95\% confidence interval (CI), where CIs are computed based on multiple runs from different randomly sampled initial conditions to capture variability in the model’s predictions. The best performance of each metric is marked in \textbf{bold}, and the second-best performance is \underline{underlined}.}
\label{tab:supp_simulation}
\begin{adjustbox}{width=\textwidth,center}
\begin{tabular}{c|c|c|c|c|c|c|c|c|c|c}
\toprule
\multicolumn{2}{c|}{Category}
& \multicolumn{1}{c|}{Ours}
& \multicolumn{7}{c}{General time series forecasting model} \\
\midrule
System
& \diagbox{Metric}{Model}
& \textbf{PhyxMamba}
& NBEATS
& NHiTS
& CrossFormer
& PatchTST
& TimesNet
& TiDE
& iTransformer \\
\midrule

\multirow{4}{*}{FFN}
& sMAPE ($\downarrow$)
& \textbf{2.16 $\pm$ 0.21}
& \underline{5.34 $\pm$ 0.35}
& 12.72 $\pm$ 1.49
& 13.00 $\pm$ 0.48
& 16.85 $\pm$ 0.91
& 11.18 $\pm$ 0.68
& 44.81 $\pm$ 2.12
& 16.98 $\pm$ 0.69 \\

& $D_{\text{frac}}$ ($\downarrow$)
& \textbf{0.003 $\pm$ 0.001}
& \underline{0.018 $\pm$ 0.004}
& 0.072 $\pm$ 0.009
& 0.075 $\pm$ 0.014
& 0.301 $\pm$ 0.019
& 0.156 $\pm$ 0.022
& 0.037 $\pm$ 0.006
& 0.112 $\pm$ 0.015 \\

& $D_{\text{stsp}}$ ($\downarrow$)
& \textbf{0.001 $\pm$ 0.000}
& \textbf{0.001 $\pm$ 0.000}
& 0.024 $\pm$ 0.002
& \textbf{0.001 $\pm$ 0.000}
& 0.132 $\pm$ 0.010
& 0.253 $\pm$ 0.011
& 0.279 $\pm$ 0.011
& 0.181 $\pm$ 0.014 \\

& $D_{\text{lyap}}$ ($\downarrow$)
& \textbf{0.001 $\pm$ 0.000}
& \underline{0.002 $\pm$ 0.000}
& \textbf{0.001 $\pm$ 0.000}
& \textbf{0.001 $\pm$ 0.000}
& \textbf{0.001 $\pm$ 0.000}
& 0.004 $\pm$ 0.000
& 0.004 $\pm$ 0.000
& 0.003 $\pm$ 0.000 \\
\midrule

\multirow{4}{*}{HR}
& sMAPE ($\downarrow$)
& \textbf{36.26 $\pm$ 6.56}
& \underline{37.86 $\pm$ 1.98}
& 68.87 $\pm$ 8.68
& 40.86 $\pm$ 3.77
& 56.87 $\pm$ 4.31
& 39.22 $\pm$ 3.01
& 124.69 $\pm$ 3.92
& 118.09 $\pm$ 4.31 \\

& $D_{\text{frac}}$ ($\downarrow$)
& \textbf{0.043 $\pm$ 0.007}
& \underline{0.047 $\pm$ 0.007}
& 0.090 $\pm$ 0.012
& 0.085 $\pm$ 0.015
& 0.195 $\pm$ 0.024
& 0.310 $\pm$ 0.028
& 0.251 $\pm$ 0.011
& 0.184 $\pm$ 0.025 \\

& $D_{\text{stsp}}$ ($\downarrow$)
& \textbf{0.001 $\pm$ 0.000}
& \textbf{0.001 $\pm$ 0.000}
& 0.007 $\pm$ 0.001
& \textbf{0.001 $\pm$ 0.000}
& 0.016 $\pm$ 0.007
& 0.391 $\pm$ 0.043
& 0.020 $\pm$ 0.006
& 0.016 $\pm$ 0.004 \\

& $D_{\text{lyap}}$ ($\downarrow$)
& \textbf{0.195 $\pm$ 0.036}
& \underline{0.215 $\pm$ 0.037}
& 0.457 $\pm$ 0.048
& 0.221 $\pm$ 0.034
& 0.440 $\pm$ 0.047
& 0.464 $\pm$ 0.057
& 0.834 $\pm$ 0.049
& 0.413 $\pm$ 0.047 \\
\bottomrule
\end{tabular}
\end{adjustbox}
\end{table*}

\begin{table*}[t]
\ContinuedFloat
\centering
\begin{adjustbox}{width=\textwidth,center}
\begin{tabular}{c|c|c|c|c|c|c|c|c|c|c|c}
\toprule
\multicolumn{2}{c|}{Category}
& \multicolumn{5}{c|}{Physics-informed dynamical system model} & \multicolumn{5}{c}{Zero-shot foundation model}\\
\midrule
System
& \diagbox{Metric}{Model}
& Koopa
& PLRNN
& nVAR
& PRC
& HoGRC
& Panda
& Chronos2
& Chronos2-S
& DynaMix
& Parrot \\
\midrule

\multirow{4}{*}{FFN}
& sMAPE ($\downarrow$)
& 29.04 $\pm$ 1.26
& 76.24 $\pm$ 6.04
& 48.31 $\pm$ 1.53
& 15.42 $\pm$ 1.15
& 14.74 $\pm$ 1.33
& 108.60 $\pm$ 2.54
& 50.81 $\pm$ 2.82
& 55.81 $\pm$ 4.25
& 118.40 $\pm$ 2.76
& 105.89 $\pm$ 1.29 \\

& $D_{\text{frac}}$ ($\downarrow$)
& 0.344 $\pm$ 0.012
& 0.079 $\pm$ 0.014
& 0.097 $\pm$ 0.007
& 0.137 $\pm$ 0.013
& 0.129 $\pm$ 0.008
& 0.220 $\pm$ 0.020
& 0.271 $\pm$ 0.016
& 0.259 $\pm$ 0.030
& 0.229 $\pm$ 0.024
& 0.056 $\pm$ 0.016 \\

& $D_{\text{stsp}}$ ($\downarrow$)
& 0.160 $\pm$ 0.010
& 0.022 $\pm$ 0.014
& 0.293 $\pm$ 0.006
& 0.231 $\pm$ 0.012
& 0.224 $\pm$ 0.009
& 0.168 $\pm$ 0.012
& 0.068 $\pm$ 0.008
& 0.278 $\pm$ 0.028
& 0.571 $\pm$ 0.235
& 0.005 $\pm$ 0.002 \\

& $D_{\text{lyap}}$ ($\downarrow$)
& \underline{0.002 $\pm$ 0.000}
& 0.007 $\pm$ 0.001
& 0.005 $\pm$ 0.000
& 0.005 $\pm$ 0.000
& 0.004 $\pm$ 0.000
& \underline{0.002 $\pm$ 0.000}
& \underline{0.002 $\pm$ 0.000}
& 0.002 $\pm$ 0.000
& \underline{0.002 $\pm$ 0.000}
& \underline{0.002 $\pm$ 0.000} \\
\midrule

\multirow{4}{*}{HR}
& sMAPE ($\downarrow$)
& 63.12 $\pm$ 5.47
& 67.42 $\pm$ 8.23
& 93.69 $\pm$ 7.67
& 74.77 $\pm$ 6.54
& 72.14 $\pm$ 5.90
& 127.99 $\pm$ 3.55
& 96.32 $\pm$ 5.78
& 98.44 $\pm$ 5.80
& 137.66 $\pm$ 6.55
& 136.85 $\pm$ 5.47 \\

& $D_{\text{frac}}$ ($\downarrow$)
& 0.227 $\pm$ 0.027
& 0.150 $\pm$ 0.015
& 0.714 $\pm$ 0.008
& 0.235 $\pm$ 0.007
& 0.227 $\pm$ 0.015
& 0.125 $\pm$ 0.027
& 0.228 $\pm$ 0.026
& 0.255 $\pm$ 0.028
& 0.203 $\pm$ 0.022
& 0.093 $\pm$ 0.019 \\

& $D_{\text{stsp}}$ ($\downarrow$)
& 0.093 $\pm$ 0.048
& 0.004 $\pm$ 0.001
& \underline{0.003 $\pm$ 0.001}
& 0.015 $\pm$ 0.003
& 0.014 $\pm$ 0.002
& 0.020 $\pm$ 0.005
& 0.014 $\pm$ 0.003
& 0.013 $\pm$ 0.003
& 0.021 $\pm$ 0.006
& 0.008 $\pm$ 0.006 \\

& $D_{\text{lyap}}$ ($\downarrow$)
& 0.423 $\pm$ 0.051
& 0.461 $\pm$ 0.056
& 1.613 $\pm$ 0.049
& 1.221 $\pm$ 0.051
& 1.175 $\pm$ 0.040
& 0.309 $\pm$ 0.041
& 0.229 $\pm$ 0.036
& 0.272 $\pm$ 0.034
& 0.451 $\pm$ 0.060
& 0.285 $\pm$ 0.051 \\
\bottomrule
\end{tabular}
\end{adjustbox}
\end{table*}

We introduce another two simulated systems, Forced FitzHugh-Nagumo and Hindmarsh-Rose, for additional evaluation. We evaluate the main representative baseline methods, and report the complete metrics in Table~\ref{tab:supp_simulation}.

\subsection{Detailed Results of Ablation Study} \label{app:ablation_results}
Table~\ref{tab:ablation_r_ecg} provides the complete numerical values and 95\% confidence intervals for the ablation study summarized in Section~\ref{sec:exp}.

\begin{table*}[t]
\centering
\caption{Model performances for Rössler  and ECG when removing each of our designs. Reported values represent the mean ± 95\% confidence interval (CI), where CIs are computed based on multiple runs from different randomly sampled initial conditions to capture variability in the model’s predictions. The best performance of each metric is marked in \textbf{bold}, and the second-best performance is \underline{underlined}.}
\label{tab:ablation_r_ecg}
\begin{adjustbox}{width=\linewidth,center}
\begin{tabular}{c|cccc|cccc}
\toprule
\midrule
System & \multicolumn{4}{c|}{Rössler} & \multicolumn{4}{c}{ECG} \\
\midrule
\multicolumn{1}{l|}{\diagbox{Ablation}{Metric}} 
& sMAPE ($\downarrow$) 
& $D_{\text{frac}}$ ($\downarrow$) 
& $D_{\text{stsp}}$ ($\downarrow$)
& $D_{\text{lyap}}$ ($\downarrow$)
& sMAPE ($\downarrow$) 
& $D_{\text{frac}}$ ($\downarrow$) 
& $D_{\text{stsp}}$ ($\downarrow$)
& $D_{\text{lyap}}$ ($\downarrow$) \\
\midrule
\textbf{Full} 
& \textbf{2.84 $\pm$ 0.18} 
& \textbf{0.050 $\pm$ 0.008} 
& \textbf{0.034 $\pm$ 0.009}
& \textbf{0.016 $\pm$ 0.004}
& \textbf{37.14 $\pm$ 15.75} 
& \textbf{0.036 $\pm$ 0.013} 
& \textbf{0.004 $\pm$ 0.001}
& \textbf{0.230 $\pm$ 0.139} \\

w/o PIE 
& 17.91$\pm$1.77 
& 0.105$\pm$0.012 
& 0.239$\pm$0.039
& 0.034 $\pm$ 0.005
& 41.02 $\pm$ 16.19 
& 0.043 $\pm$ 0.013 
& \underline{0.006 $\pm$ 0.002}
& 0.406 $\pm$ 0.150 \\

w/o RS 
& 11.29$\pm$1.01 
& \underline{0.075$\pm$0.010} 
& 0.150$\pm$0.028
& \underline{0.026 ± 0.004}
& 41.05 $\pm$ 15.54 
& \underline{0.044 $\pm$ 0.012} 
& 0.011 $\pm$ 0.003
& 0.380 $\pm$ 0.152 \\

w/o MPG 
& 12.48$\pm$1.21 
& 0.091$\pm$0.012 
& 0.193$\pm$0.039
& 0.030 $\pm$ 0.004
& \underline{40.20 $\pm$ 15.73} 
& 0.043 $\pm$ 0.013 
& 0.010 $\pm$ 0.004
& 0.370 $\pm$ 0.126 \\

w/o SF 
& 27.22$\pm$0.83 
& 0.108$\pm$0.012 
& 0.230$\pm$0.041
& 0.037 $\pm$ 0.005
& 43.80 $\pm$ 16.08 
& 0.042 $\pm$ 0.016 
& 0.012 $\pm$ 0.005
& \underline{0.302 $\pm$ 0.123} \\

w/o MMD 
& \underline{8.34$\pm$0.72} 
& 0.083$\pm$0.012 
& \underline{0.136$\pm$0.024}
& 0.028 ± 0.004
& 49.29 $\pm$ 18.19 
& 0.269 $\pm$ 0.045 
& 0.099 $\pm$ 0.027
& 0.351 $\pm$ 0.116 \\

Encoder
& 20.68$\pm$1.47 
& 0.081$\pm$0.011 
& 0.160$\pm$0.031
& 0.035 $\pm$ 0.004
& 53.48 $\pm$ 17.48 
& 0.905 $\pm$ 0.082 
& 0.200 $\pm$ 0.035
& 0.353 $\pm$ 0.120 \\
\midrule
\bottomrule
\end{tabular}
\end{adjustbox}
\end{table*}

\subsection{Detailed Results of Model Robustness} \label{app:app_robustness}
Tables~\ref{tab:noise} and~\ref{tab:ratio} report the full numerical results with 95\% confidence intervals for the noise-intensity and training-data-ratio analyses summarized in Section~\ref{sec:exp}.

\begin{table*}[t]
\centering
\caption{Robustness under different noise intensities.}
\label{tab:noise}
\begin{adjustbox}{width=\textwidth,center}
\begin{tabular}{c|cc|cc|cc|cc}
\toprule
\midrule
\multicolumn{1}{c|}{System} 
& \multicolumn{2}{c|}{Lorenz63} 
& \multicolumn{2}{c|}{Rössler} 
& \multicolumn{2}{c|}{Lorenz96} 
& \multicolumn{2}{c}{ECG} \\
\midrule
\multicolumn{1}{l|}{\diagbox{Intensity}{Metric}} &
\multicolumn{1}{c}{sMAPE ($\downarrow$)} & \multicolumn{1}{c|}{$D_{\text{stsp}}$ ($\downarrow$)} &
\multicolumn{1}{c}{sMAPE ($\downarrow$)} & \multicolumn{1}{c|}{$D_{\text{stsp}}$ ($\downarrow$)} &
\multicolumn{1}{c}{sMAPE ($\downarrow$)} & \multicolumn{1}{c|}{$D_{\text{stsp}}$ ($\downarrow$)} &
\multicolumn{1}{c}{sMAPE ($\downarrow$)} & \multicolumn{1}{c}{$D_{\text{stsp}}$ ($\downarrow$)} \\
\midrule
0.0000 & 3.26 $\pm$ 0.56 & 1.133 $\pm$ 0.340 & 2.85 $\pm$ 0.17 & 0.034 $\pm$ 0.009 & 20.28 $\pm$ 2.69 & 15.440 $\pm$ 0.799 & 37.14 $\pm$ 15.75 & 0.004 $\pm$ 0.001 \\
0.0001 & 3.33 $\pm$ 0.72 & 1.272 $\pm$ 0.345 & 2.21 $\pm$ 0.22 & 0.049 $\pm$ 0.012 & 19.70 $\pm$ 2.49 & 15.496 $\pm$ 0.719 & 38.00 $\pm$ 13.40 & 0.003 $\pm$ 0.003 \\
0.0002 & 3.09 $\pm$ 0.58 & 0.992 $\pm$ 0.246 & 3.49 $\pm$ 0.26 & 0.064 $\pm$ 0.016 & 19.59 $\pm$ 2.52 & 14.884 $\pm$ 0.670 & 37.01 $\pm$ 12.94 & 0.006 $\pm$ 0.002 \\
0.0005 & 3.49 $\pm$ 0.83 & 1.166 $\pm$ 0.324 & 2.79 $\pm$ 0.27 & 0.054 $\pm$ 0.014 & 19.35 $\pm$ 2.49 & 15.385 $\pm$ 0.650 & 39.04 $\pm$ 14.21 & 0.005 $\pm$ 0.002 \\
0.0010 & 3.39 $\pm$ 0.64 & 1.006 $\pm$ 0.301 & 3.45 $\pm$ 0.27 & 0.063 $\pm$ 0.017 & 19.48 $\pm$ 2.54 & 15.948 $\pm$ 0.841 & 38.80 $\pm$ 13.59 & 0.006 $\pm$ 0.002 \\
0.0020 & 3.86 $\pm$ 0.88 & 0.956 $\pm$ 0.255 & 3.84 $\pm$ 0.26 & 0.057 $\pm$ 0.021 & 18.78 $\pm$ 2.47 & 15.571 $\pm$ 0.700 & 38.90 $\pm$ 13.42 & 0.010 $\pm$ 0.004 \\
0.0050 & 3.64 $\pm$ 1.00 & 1.111 $\pm$ 0.253 & 5.43 $\pm$ 0.29 & 0.056 $\pm$ 0.017 & 19.01 $\pm$ 2.51 & 15.198 $\pm$ 0.663 & 37.83 $\pm$ 13.36 & 0.011 $\pm$ 0.003 \\
0.0100 & 3.00 $\pm$ 0.64 & 0.898 $\pm$ 0.255 & 9.30 $\pm$ 0.43 & 0.067 $\pm$ 0.017 & 19.96 $\pm$ 2.44 & 16.187 $\pm$ 0.824 & 38.60 $\pm$ 13.40 & 0.005 $\pm$ 0.002 \\
0.0200 & 3.37 $\pm$ 1.03 & 1.093 $\pm$ 0.261 & 27.97 $\pm$ 0.89 & 0.082 $\pm$ 0.029 & 18.62 $\pm$ 2.34 & 15.370 $\pm$ 0.850 & 46.06 $\pm$ 14.21 & 0.011 $\pm$ 0.002 \\
0.0500 & 5.24 $\pm$ 1.35 & 1.238 $\pm$ 0.259 & 37.80 $\pm$ 1.05 & 0.095 $\pm$ 0.015 & 19.05 $\pm$ 2.27 & 15.381 $\pm$ 0.742 & 64.42 $\pm$ 17.91 & 0.009 $\pm$ 0.004 \\
0.1000 & 8.34 $\pm$ 2.08 & 1.152 $\pm$ 0.248 & 44.63 $\pm$ 1.63 & 0.110 $\pm$ 0.018 & 21.89 $\pm$ 2.56 & 15.466 $\pm$ 0.756 & 80.75 $\pm$ 15.92 & 0.011 $\pm$ 0.003 \\
0.2000 & 8.58 $\pm$ 2.28 & 1.162 $\pm$ 0.315 & 45.34 $\pm$ 1.63 & 0.108 $\pm$ 0.018 & 25.95 $\pm$ 2.87 & 15.742 $\pm$ 0.915 & 118.70 $\pm$ 9.06 & 0.047 $\pm$ 0.006 \\
0.5000 & 14.73 $\pm$ 2.11 & 1.559 $\pm$ 0.495 & 55.86 $\pm$ 1.72 & 0.168 $\pm$ 0.020 & 36.33 $\pm$ 3.24 & 16.497 $\pm$ 0.669 & 119.26 $\pm$ 9.06 & 0.048 $\pm$ 0.006 \\
\midrule
\bottomrule
\end{tabular}
\end{adjustbox}
\end{table*}

\begin{table*}[t]
\centering
\caption{Robustness under different training-data ratios.}
\label{tab:ratio}
\begin{adjustbox}{width=\textwidth,center}
\begin{tabular}{c|cc|cc|cc|cc}
\toprule
\midrule
\multicolumn{1}{c|}{System} &
\multicolumn{2}{c|}{Lorenz63} &
\multicolumn{2}{c|}{Rössler} &
\multicolumn{2}{c|}{Lorenz96} &
\multicolumn{2}{c}{ECG} \\
\midrule
\multicolumn{1}{l|}{\diagbox{Ratio}{Metric}} &
\multicolumn{1}{c}{sMAPE ($\downarrow$)} & \multicolumn{1}{c|}{$D_{\text{stsp}}$ ($\downarrow$)} &
\multicolumn{1}{c}{sMAPE ($\downarrow$)} & \multicolumn{1}{c|}{$D_{\text{stsp}}$ ($\downarrow$)} &
\multicolumn{1}{c}{sMAPE ($\downarrow$)} & \multicolumn{1}{c|}{$D_{\text{stsp}}$ ($\downarrow$)} &
\multicolumn{1}{c}{sMAPE ($\downarrow$)} & \multicolumn{1}{c}{$D_{\text{stsp}}$ ($\downarrow$)} \\
\midrule
0.01 &
\multicolumn{1}{c|}{18.13 $\pm$ 2.66} & 1.189 $\pm$ 0.197 &
\multicolumn{1}{c|}{20.62 $\pm$ 0.99} & 0.212 $\pm$ 0.037 &
\multicolumn{1}{c|}{65.62 $\pm$ 4.53} & 17.518 $\pm$ 0.711 &
\multicolumn{1}{c|}{43.32 $\pm$ 16.31} & 0.018 $\pm$ 0.005 \\

0.05 &
\multicolumn{1}{c|}{7.31 $\pm$ 1.12} & 1.065 $\pm$ 0.332 &
\multicolumn{1}{c|}{16.91 $\pm$ 0.95} & 0.204 $\pm$ 0.037 &
\multicolumn{1}{c|}{36.49 $\pm$ 4.15} & 16.546 $\pm$ 0.762 &
\multicolumn{1}{c|}{37.90 $\pm$ 10.55} & 0.005 $\pm$ 0.002 \\

0.10 &
\multicolumn{1}{c|}{6.20 $\pm$ 0.94} & 1.043 $\pm$ 0.239 &
\multicolumn{1}{c|}{13.22 $\pm$ 0.98} & 0.183 $\pm$ 0.037 &
\multicolumn{1}{c|}{28.64 $\pm$ 3.70} & 15.659 $\pm$ 0.748 &
\multicolumn{1}{c|}{37.54 $\pm$ 11.80} & 0.005 $\pm$ 0.003 \\

0.15 &
\multicolumn{1}{c|}{4.78 $\pm$ 0.81} & 1.412 $\pm$ 0.523 &
\multicolumn{1}{c|}{10.16 $\pm$ 0.84} & 0.173 $\pm$ 0.035 &
\multicolumn{1}{c|}{27.19 $\pm$ 3.58} & 16.264 $\pm$ 0.801 &
\multicolumn{1}{c|}{37.70 $\pm$ 10.75} & 0.004 $\pm$ 0.002 \\

0.20 &
\multicolumn{1}{c|}{4.09 $\pm$ 0.61} & 1.105 $\pm$ 0.236 &
\multicolumn{1}{c|}{9.85 $\pm$ 0.57} & 0.109 $\pm$ 0.017 &
\multicolumn{1}{c|}{23.30 $\pm$ 3.11} & 15.791 $\pm$ 0.676 &
\multicolumn{1}{c|}{36.66 $\pm$ 11.05} & 0.007 $\pm$ 0.003 \\

0.25 &
\multicolumn{1}{c|}{3.61 $\pm$ 0.58} & 1.170 $\pm$ 0.316 &
\multicolumn{1}{c|}{10.72 $\pm$ 0.91} & 0.168 $\pm$ 0.034 &
\multicolumn{1}{c|}{22.25 $\pm$ 2.76} & 15.186 $\pm$ 0.711 &
\multicolumn{1}{c|}{38.33 $\pm$ 11.56} & 0.007 $\pm$ 0.003 \\

0.40 &
\multicolumn{1}{c|}{3.41 $\pm$ 0.59} & 1.271 $\pm$ 0.438 &
\multicolumn{1}{c|}{5.23 $\pm$ 0.47} & 0.092 $\pm$ 0.020 &
\multicolumn{1}{c|}{19.62 $\pm$ 2.42} & 15.244 $\pm$ 0.710 &
\multicolumn{1}{c|}{38.35 $\pm$ 12.55} & 0.004 $\pm$ 0.002 \\

0.60 &
\multicolumn{1}{c|}{3.18 $\pm$ 0.56} & 1.110 $\pm$ 0.266 &
\multicolumn{1}{c|}{4.93 $\pm$ 0.34} & 0.072 $\pm$ 0.014 &
\multicolumn{1}{c|}{19.92 $\pm$ 2.32} & 15.465 $\pm$ 0.778 &
\multicolumn{1}{c|}{38.28 $\pm$ 13.07} & 0.004 $\pm$ 0.002 \\

0.80 &
\multicolumn{1}{c|}{3.44 $\pm$ 0.76} & 1.321 $\pm$ 0.530 &
\multicolumn{1}{c|}{5.12 $\pm$ 0.49} & 0.085 $\pm$ 0.015 &
\multicolumn{1}{c|}{20.66 $\pm$ 2.53} & 15.524 $\pm$ 0.834 &
\multicolumn{1}{c|}{37.58 $\pm$ 13.50} & 0.004 $\pm$ 0.002 \\

1.00 &
\multicolumn{1}{c|}{3.26 $\pm$ 0.56} & 1.133 $\pm$ 0.340 &
\multicolumn{1}{c|}{2.84 $\pm$ 0.18} & 0.034 $\pm$ 0.009 &
\multicolumn{1}{c|}{20.28 $\pm$ 2.69} & 15.440 $\pm$ 0.799 &
\multicolumn{1}{c|}{37.14 $\pm$ 15.75} & 0.004 $\pm$ 0.001 \\
\midrule
\bottomrule
\end{tabular}
\end{adjustbox}
\end{table*}

\subsection{Hyperparameter Study}\label{app:hyperparameter-study}
We provide supplementary hyperparameter analyses for the time-delay embedding parameters, latent patch dimension, patch size, multi-patch evolutionary guidance steps, model depth, and the guidance weight $\lambda_p$. Unless otherwise stated, appendix tables report mean $\pm$ 95\% confidence intervals across multiple runs. The results are reported in Tables~\ref{tab:impact_m_lorenz63},~\ref{tab:impact_tau_lorenz63},~\ref{tab:impact_d_lorenz63},~\ref{tab:patch_with_ci_l63},~\ref{tab:mtp_with_ci_l63},~\ref{tab:hier_with_ci_l63}, and~\ref{tab:impact_lambda_p_lorenz63}.

\noindent \textbf{Impact of Time-delay Embedding Parameters.} The embedding dimension $m$ and time delay $\tau$ determine how the observed trajectory is unfolded into a reconstructed phase space. As shown in Tables~\ref{tab:impact_m_lorenz63} and~\ref{tab:impact_tau_lorenz63}, suboptimal choices can degrade performance because the delayed coordinates may fail to properly unfold the attractor. In practice, this issue is mitigated by the data-driven selection rules in Section~\ref{sec:method-rep}, where $\tau$ is chosen by average mutual information and $m$ is selected after the reconstructed neighborhood statistics stabilize.

\begin{table*}[!t]
\centering
\begin{minipage}[t]{0.48\textwidth}
\centering
\caption{Effect of embedding dimension $m$ on Lorenz63.}
\label{tab:impact_m_lorenz63}
{\scriptsize
\setlength{\tabcolsep}{3pt}
\renewcommand{\arraystretch}{0.9}
\begin{tabular}{@{}ccc@{}}
\toprule
\textbf{$m$} & sMAPE ($\downarrow$) & $D_\text{stsp}$ ($\downarrow$) \\ \midrule
1 & 3.51 $\pm$ 1.06 & 0.989 $\pm$ 0.244 \\
2 & 3.52 $\pm$ 0.86 & 1.099 $\pm$ 0.241 \\
3 & 3.26 $\pm$ 0.56 & 1.133 $\pm$ 0.340 \\
4 & 3.54 $\pm$ 0.66 & 1.273 $\pm$ 0.404 \\
5 & 5.92 $\pm$ 2.26 & 1.006 $\pm$ 0.243 \\
7 & 4.64 $\pm$ 0.80 & 0.912 $\pm$ 0.202 \\
9 & 3.66 $\pm$ 0.66 & 0.924 $\pm$ 0.168 \\ \bottomrule
\end{tabular}
}
\end{minipage}
\hfill
\begin{minipage}[t]{0.48\textwidth}
\centering
\caption{Effect of time delay $\tau$ on Lorenz63.}
\label{tab:impact_tau_lorenz63}
{\scriptsize
\setlength{\tabcolsep}{3pt}
\renewcommand{\arraystretch}{0.9}
\begin{tabular}{@{}ccc@{}}
\toprule
$\tau$ & sMAPE ($\downarrow$) & $D_{\text{stsp}}$ ($\downarrow$) \\ \midrule
1 & 4.81 $\pm$ 1.92 & 1.100 $\pm$ 0.273 \\
2 & 4.05 $\pm$ 1.15 & 0.970 $\pm$ 0.188 \\
3 & 3.41 $\pm$ 0.91 & 1.154 $\pm$ 0.250 \\
4 & 3.52 $\pm$ 0.85 & 1.212 $\pm$ 0.385 \\
5 & 3.58 $\pm$ 0.60 & 1.130 $\pm$ 0.270 \\
7 & 3.26 $\pm$ 0.56 & 1.133 $\pm$ 0.340 \\
9 & 3.54 $\pm$ 0.59 & 1.185 $\pm$ 0.310 \\ \bottomrule
\end{tabular}
}
\end{minipage}
\end{table*}

\noindent \textbf{Impact of Patch Size.} Patch size plays a crucial role in determining representation quality, particularly in time series data where patches often capture local semantic patterns. An appropriately chosen patch size can enhance the model’s ability to extract meaningful temporal features, whereas suboptimal sizing may either omit important short-term dynamics or dilute significant signals. Therefore, tailoring the patch size to the specific characteristics of the target system is essential for optimal performance.

\noindent \textbf{Impact of Multi-patch Evolutionary Guidance Steps.} The number of multi-patch evolutionary guidance steps also affects the model’s capacity to learn temporal dependencies. A step size that is too small limits the model’s ability to capture the long-term dynamics necessary for accurate forecasting, while an excessively large step size may hinder next-patch prediction performance and increase computational complexity. Hence, it is imperative to strike a balance: the chosen multi-patch evolutionary guidance steps should be sufficiently large to reflect system-level temporal evolution, yet not so large as to compromise local accuracy or computational efficiency.

\noindent Tables~\ref{tab:impact_d_lorenz63} and~\ref{tab:hier_with_ci_l63} provide the complete confidence intervals for the latent-dimension and Mamba-layer analyses discussed in Section~\ref{sec:hyperparameter-study}.

\noindent \textbf{Impact of $\lambda_p$ values.} We conduct an additional experiment on the Lorenz63 dataset to investigate the effect of $\lambda_p$. The results are shown in Table~\ref{tab:impact_lambda_p_lorenz63}. $\lambda_p$ controls the balance between next-patch prediction and multi-patch evolutionary guidance, and improper values may weaken either short-term accuracy or long-term dynamical fidelity.

\begin{table*}[!t]
\centering
\begin{minipage}[t]{0.48\textwidth}
\centering
\caption{Effect of patch size $D$ on Lorenz63.}
\label{tab:patch_with_ci_l63}
{\scriptsize
\setlength{\tabcolsep}{4pt}
\renewcommand{\arraystretch}{0.88}
\begin{tabular}{ccc}
\toprule
$D$ & sMAPE & $D_{\text{stsp}}$ \\
\midrule
2  & $6.14 \pm 1.30$ & $0.992 \pm 0.181$ \\
3  & $4.64 \pm 0.84$ & $1.275 \pm 0.366$ \\
5  & $3.60 \pm 0.46$ & $1.399 \pm 0.286$ \\
6  & $3.74 \pm 0.99$ & $1.093 \pm 0.328$ \\
10 & $3.26 \pm 0.56$ & $1.133 \pm 0.340$ \\
\bottomrule
\end{tabular}
}
\par\smallskip

\caption{Effect of MPG steps $M$ on Lorenz63.}
\label{tab:mtp_with_ci_l63}
{\scriptsize
\setlength{\tabcolsep}{4pt}
\renewcommand{\arraystretch}{0.88}
\begin{tabular}{ccc}
\toprule
$M$ & sMAPE & $D_{\text{stsp}}$ \\
\midrule
1 & $3.07 \pm 0.52$ & $1.293 \pm 0.235$ \\
2 & $2.98 \pm 0.56$ & $1.420 \pm 0.442$ \\
3 & $3.26 \pm 0.56$ & $1.133 \pm 0.340$ \\
4 & $4.12 \pm 0.76$ & $1.212 \pm 0.358$ \\
\bottomrule
\end{tabular}
}
\par\smallskip

\caption{Effect of latent dimension $d$ on Lorenz63.}
\label{tab:impact_d_lorenz63}
{\scriptsize
\setlength{\tabcolsep}{4pt}
\renewcommand{\arraystretch}{0.88}
\begin{tabular}{@{}ccc@{}}
\toprule
$d$ & sMAPE ($\downarrow$) & $D_\text{stsp}$ ($\downarrow$) \\ \midrule
16  & 21.25 $\pm$ 3.52 & 1.409 $\pm$ 0.353 \\
32  & 14.82 $\pm$ 2.43 & 1.225 $\pm$ 0.241 \\
64  & 15.27 $\pm$ 3.27 & 1.279 $\pm$ 0.317 \\
128 & 5.51 $\pm$ 1.35  & 1.074 $\pm$ 0.345 \\
256 & 3.26 $\pm$ 0.56  & 1.133 $\pm$ 0.340 \\
512 & 1.49 $\pm$ 0.26  & 1.051 $\pm$ 0.200 \\ \bottomrule
\end{tabular}
}
\end{minipage}
\hfill
\begin{minipage}[t]{0.48\textwidth}
\centering
\caption{Effect of Mamba layers $L$ on Lorenz63.}
\label{tab:hier_with_ci_l63}
{\scriptsize
\setlength{\tabcolsep}{4pt}
\renewcommand{\arraystretch}{0.88}
\begin{tabular}{ccc}
\toprule
$L$ & sMAPE & $D_{\text{stsp}}$ \\
\midrule
1 & $3.90 \pm 0.76$ & $1.331 \pm 0.335$ \\
2 & $3.65 \pm 0.40$ & $1.231 \pm 0.263$ \\
3 & $3.60 \pm 0.46$ & $1.199 \pm 0.286$ \\
4 & $3.26 \pm 0.56$ & $1.133 \pm 0.340$ \\
5 & $4.25 \pm 0.94$ & $1.176 \pm 0.298$ \\
\bottomrule
\end{tabular}
}
\par\smallskip

\caption{Effect of guidance weight $\lambda_p$ on Lorenz63.}
\label{tab:impact_lambda_p_lorenz63}
{\scriptsize
\setlength{\tabcolsep}{4pt}
\renewcommand{\arraystretch}{0.88}
\begin{tabular}{@{}ccc@{}}
\toprule
$\lambda_p$ & sMAPE ($\downarrow$) & $D_\text{stsp}$ ($\downarrow$) \\ \midrule
0.001 & 2.96 $\pm$ 0.67 & 1.265 $\pm$ 0.365 \\
0.01  & 3.71 $\pm$ 0.87 & 1.287 $\pm$ 0.243 \\
0.1   & 3.26 $\pm$ 0.56 & 1.133 $\pm$ 0.340 \\
1     & 3.28 $\pm$ 0.49 & 1.213 $\pm$ 0.360 \\
10    & 3.41 $\pm$ 0.46 & 1.231 $\pm$ 0.355 \\ \bottomrule
\end{tabular}
}
\end{minipage}
\end{table*}

\end{document}